\documentclass{article}
\usepackage{subcaption}
\usepackage{microtype}
\usepackage{graphicx}
\usepackage{subcaption}
\usepackage{booktabs} 
\usepackage{colortbl}
\usepackage{xcolor}
\usepackage{hyperref}

\usepackage[accepted]{icml2025}

\usepackage{amsmath}
\usepackage{amssymb}
\usepackage{mathtools}
\usepackage{amsthm}
\usepackage{wrapfig}
\usepackage[capitalize,noabbrev]{cleveref}

\theoremstyle{plain}

\theoremstyle{definition}

\theoremstyle{remark}

\usepackage[textsize=tiny]{todonotes}

\icmltitlerunning{Submission and Formatting Instructions for ICML 2025}

\begin{document}

\twocolumn[
\icmltitle{Long Context Scaling: Divide and Conquer via \\Multi-Agent Question-driven Collaboration}



\icmlsetsymbol{equal}{*}

\begin{icmlauthorlist}
\icmlauthor{Sibo Xiao}{zju}
\icmlauthor{Zixin Lin}{zju}
\icmlauthor{Wenyang Gao}{zju,wlu}
\icmlauthor{Hui Chen}{zju}
\icmlauthor{Yue Zhang}{wlu}
\end{icmlauthorlist}

\icmlaffiliation{zju}{College of Computer Science, Zhejiang University, E-mails: \texttt{sibodotxiao@gmail.com}}
\icmlaffiliation{wlu}{NLP Lab, Westlake University}

\icmlcorrespondingauthor{Yue Zhang}{zhangyue@westlake.edu.cn}

\icmlkeywords{Machine Learning, ICML}

\vskip 0.3in
]



\printAffiliationsAndNotice{} 

\begin{abstract}
Processing long contexts has become a critical capability for modern large language models (LLMs). Existing works leverage agent-based divide-and-conquer methods for processing long contexts. But these methods face crucial limitations, including prohibitive accumulated latency and amplified information loss from excessive agent invocations, and the disruption of inherent textual dependencies by immoderate partitioning. XpandA tackles long-text challenges using: 1) dynamic partitioning for flexible context handling; 2) question-driven updates for consistent knowledge across partitions; and 3) selective replay of partitions to resolve complex information order (e.g., flashbacks). We perform a comprehensive evaluation of XpandA on multiple long-context benchmarks with length varying from 1k to 1M, demonstrating XpandA's feasibility for processing ultra-long sequences and its significant effectiveness in enhancing the long-context capabilities of various LLMs by achieving 20\% improvements and 1.5x inference speedup over baselines of full-context, RAG and previous agent-based methods.
\end{abstract}

\section{Introduction}
Like human cognition—which depends on long-term memory, accumulated knowledge, and life experience—Artificial General Intelligence (AGI) development requires the ability to associate distant information and retain context over extended content \cite{lake2016buildingmachineslearnthink,morris2024levelsagioperationalizingprogress}. Long-context tasks, such as question answering \cite{wang2025novelqabenchmarkingquestionanswering, yang-etal-2018-hotpotqa, narrativeqa}, summarization \cite{zhong-etal-2021-qmsum} and coding \cite{bogomolov2024longcodearenaset}, consistently present significant challenges due to issues like hallucination and overlooking critical information \cite{liu2023lostmiddlelanguagemodels}.

Recent efforts to tackle these challenges have focused on three key directions: training data strategy, architecture and workflow design. The training data strategy conducts long-context data filtering \cite{wu_longattn_2025,si_gateau_2025}, mixture \cite{fu2024dataengineeringscalinglanguage,jin2023growlengthacceleratingllmspretraining}, and synthesis \cite{gao2025questquerycentricdatasynthesis,he-etal-2024-never} on pre-training and post-training phase. Architecture methods employ position embeddings  \cite{Su2021RoFormerET, Press2021TrainST, Golovneva2024ContextualPE, Sun2022ALT} and attention mechanisms such as transformer-based \cite{beltagy_longformer_2020, xiao_duoattention_2024, zhou_progressive_2025,yuan_native_2025, fu2024moamixturesparseattention}, linear-complexity \cite{qin_lightning_2024} and hybrid architectures \cite{munkhdalai_leave_2024, lieber2024jambahybridtransformermambalanguage, lu2025mobamixtureblockattention}. Despite these efforts, the lack of extensibility in context windows leads to input truncation when processing longer sequences, while training long-context language models entails prohibitively high computational costs\cite{beltagy_longformer_2020}. Workflow design, including external memory modules \cite{wang2023augmentinglanguagemodelslongterm,mehta2023dsiupdatingtransformermemory}, retrieval-augmented generation (RAG) \cite{borgeaud2022improvinglanguagemodelsretrieving, joshi2024reaperreasoningbasedretrieval}, and agent-based methods (LongAgent \cite{zhao2024longagentscalinglanguagemodels}, Chain of Agents \cite{zhang2024chainagentslargelanguage}), etc., can mitigate the challenges mentioned above but introduce issues such as information loss for inappropriate extraction \cite{zhang2024chainagentslargelanguage}, Inter-Agent Misalignment and Specification Violation inherent in the multi-agent system \cite{cemri2025multiagentllmsystemsfail}. Furthermore, the most concerning aspect of agent-based methods is the frequent model invocation. This not only potentially amplifies intermediate errors but also leads to an accumulation of model call latency. 

Motivated by the aforementioned challenges, we propose a multi-agent framework XpandA (Expand-Agent) that incorporates fine-grained question-driven communication protocols and, for the first time, extends the input sequence length of agent-based methods beyond 128k tokens. By introducing a Saturating Function within its partitioning strategy, XpandA can practically process input sequences with vastly varying lengths and achieve a trade-off between unit inference time and accumulated model call latency. Furthermore, XpandA effectively guides the model's attention transfer across partitions via its question-driven communication protocol, thereby mitigating information dependency fragmentation caused by partitioning. Inspired by Activity On Vertex (AOV) networks, we implement an iterative selective replay mechanism based on collected information-question couples. This approach effectively resolves potential topological structures of information across different text partitions (such as narratives with flashbacks or interpolations that cannot be resolved sequentially).
\begin{figure*}[t]
  \centering
  \includegraphics[width=\textwidth]{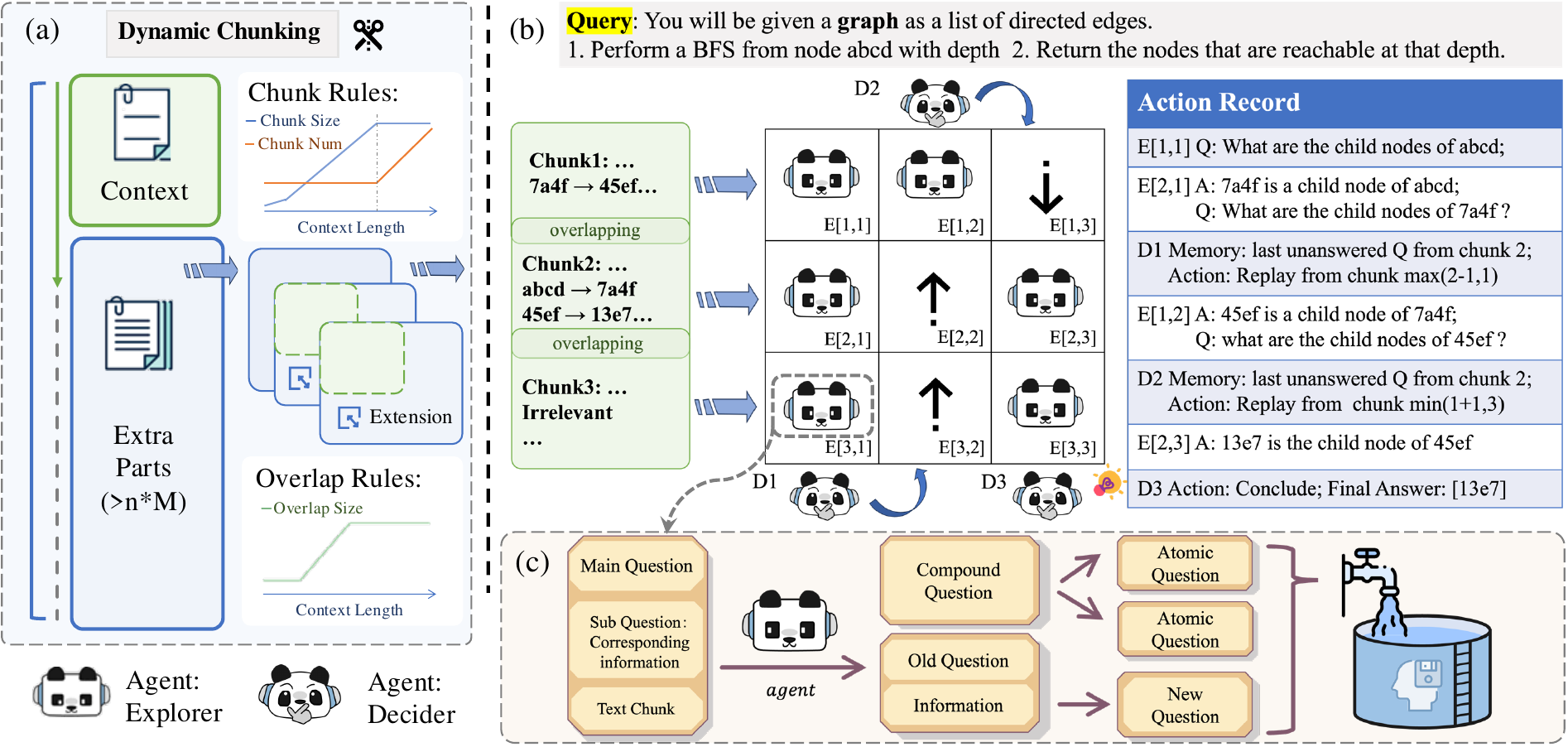}
  \caption{An Overview of XpandA. A plug-and-play, robust, interpretable, and efficient multi-agent system framework that introduces a dynamic chunking strategy (a) and selective replay (b) in its question-driven workflow (c). XpandA accepts long context input and splits the input into chunks with overlap, and each chunk is fed to corresponding Explorer Agents $\{E[i,1]\ |\ i\in{1,2,3}\}$ sequentially. Then Decider D1 decides the action of the next step is to Replay or Conclude. If Replay is proposed, {$E[i,2]$} will be called in a reverse direction with a start point according to the question state tracker.}
  \label{fig:overview}
\end{figure*}
We conduct comprehensive experiments on long-context benchmarks with diverse tasks of vastly varying lengths (from 1k to 1M). Utilizing open-source LLMs, we compare XpandA with strong baselines of full-context, Retrieval-Augmented Generation (RAG), and prior Multi-agent methods, demonstrating its significant improvement over all baselines by up to 20\% and 1.5x inference speedup. In summary, our contributions to the research community of long-context processing include:
\begin{enumerate}
\item Proposing a multi-agent framework with fine-grained question-driven communication protocols that can practically expand the LLM's context window beyond 128k without training. 

\item Developing a dynamic partitioning strategy that achieves asymptotically optimal inference speedup while enhancing inference performance by trade-off between unit inference time and accumulated model call latency with input sequences of vastly varying lengths (from 1k to 1M).

\item Establishing a paradigm for multi-agent system in long-context tasks domain to mitigate inter-agent misalignment and specification violations using principled prompt and structured workflow.
\end{enumerate}

\section{Related Work}
\subsection{Long Context Language Modeling}
Long context processing in LLMs has recently become a central focus \cite{Liu2025ACS}, driven by tasks such as long-document QA \cite{Kocisk2017TheNR, Dasigi2021ADO,wang2025novelqabenchmarkingquestionanswering,masry_longfin_2024,fan_medodyssey_2024}, summarization \cite{huang-etal-2021-efficient, zhong-etal-2021-qmsum}, retrieval-augmented generation (RAG) \cite{Lee2024CanLL}, in-context learning \cite{Li2024LongcontextLS, Xu2024StressTestingLL}, and large-codebase understanding \cite{jimenez2024swebench,bogomolov2024longcodearenaset}. These require models to locate, aggregate, and reason over dispersed information, while overcoming \textbf{"lost in the middle"}\cite{liu2023lostmiddlelanguagemodels} issues. Existing approaches fall into three main categories: Training data strategies, architectural and workflow design \cite{liu2025comprehensivesurveylongcontext}. Data strategies enhance a model's capacity to process lengthy texts through the filter \cite{wu_longattn_2025,si_gateau_2025}, mixture \cite{fu2024dataengineeringscalinglanguage,jin2023growlengthacceleratingllmspretraining}, and synthesis \cite{gao2025questquerycentricdatasynthesis,he-etal-2024-never} of long-context data in both pre-training and post-training phases. 
Architecture design aims to develop model structures capable of effectively handling extended texts by position embeddings \cite{Su2021RoFormerET, Press2021TrainST, Golovneva2024ContextualPE, Sun2022ALT} and attention mechanisms such as transformer-based \cite{beltagy_longformer_2020, xiao_duoattention_2024, zhou_progressive_2025,yuan_native_2025, fu2024moamixturesparseattention}, linear-complexity \cite{qin_lightning_2024} and hybrid architectures \cite{munkhdalai_leave_2024, lieber2024jambahybridtransformermambalanguage, lu2025mobamixtureblockattention}. Workflow-based methods minimize model changes, using techniques like external memory \cite{wang2023augmentinglanguagemodelslongterm,mehta2023dsiupdatingtransformermemory}, RAG \cite{borgeaud2022improvinglanguagemodelsretrieving, joshi2024reaperreasoningbasedretrieval}, or agent-based reasoning \cite{chen2023walkingmemorymazecontext,li2024graphreaderbuildinggraphbasedagent,zhang2024chainagentslargelanguage}, to further improve the long context capability. 


\subsection{LLM Reasoning with Search and Planning}

Human problem-solving relies on decomposing complex tasks into manageable steps \cite{theoryps2015}, a paradigm successfully adapted to LLMs. Early work by \cite{wei2022chain} and \cite{kojima2022large} introduced Chain-of-Thought (CoT) prompting, using intermediate steps to guide reasoning, while later studies like \cite{dua-etal-2022-successive} and \cite{zhou2023leasttomost} proposed iterative decomposition into subtasks. To enhance reasoning structures, \cite{zhou2024selfdiscover} and \cite{aswani-etal-2024-auto} developed frameworks for self-discovering and dynamically adapting reasoning modules. Further, methods like \cite{deng-etal-2024-towards} and \cite{zhang-etal-2024-small} address error correction in intermediate steps, particularly for smaller models. Beyond linear reasoning, exploration-based approaches leverage search over multiple paths, inspired by human problem-solving \cite{stanovich2000individual}. While early methods like self-consistency \cite{wang2023selfconsistency} generated limited diversity, advanced frameworks such as Tree-of-Thoughts \cite{yao2023tree} and Graph-of-Thoughts \cite{besta2024graph} enable finer-grained branching. Aggregation strategies further improve robustness, ranging from ensemble-based voting \cite{wang2024soft,li2024more} to verifier-guided selection \cite{wang2024chain}.


\subsection{Multi-Agent LLMs}

Research efforts in Multi-Agent Systems have encompassed diverse aspects of collaboration, covering mechanisms like cooperation \cite{chen2024agentverse, shinn2023reflexion}, competition \cite{chen-etal-2024-llmarena, zhao2024competeai}, and coopetition \cite{abdelnabi2024cooperation, davidson2024evaluating}, as well as strategies such as rule-based \cite{zhang-etal-2024-exploring, Zhuang2024PoSE}, role-based \cite{chen2024agentverse, hong2024metagpt}, and model-based approaches \cite{li-etal-2023-theory, xu2023magic}. Furthermore, other key considerations include communication structures (centralized \cite{jiang-etal-2023-llm, ning2024skeletonofthought}, decentralized \cite{chen2024agentverse, yin-etal-2023-exchange}, and hierarchical \cite{li2023camel, liu2024a}), which offer scenario-dependent advantages and disadvantages, and coordination architectures (static or dynamic \cite{chen-etal-2024-comm, jeyakumar2024advancing, wang-etal-2024-unleashing}), providing trade-offs between stability and flexibility. MASs demonstrate broad potential for applications across numerous domains, including 5G/6G networks \cite{wang2024largelanguagemodelenabled, 10531769}, Natural Language Generation (NLG) and Question Answering (QA) \cite{jiang-etal-2023-llm, suzgun2024metapromptingenhancinglanguagemodels, wu2024autogen} and socio-cultural simulations \cite{li2024cultureparkboostingcrossculturalunderstanding, 10.1145/3627673.3679768}, among others. Nevertheless, the field still confronts numerous challenges \cite{Cemri2025WhyDM}. These include achieving unified governance, understanding the inherent limitations of individual agents \cite{zhang-etal-2024-exploring, shayegani2023survey}, scalability, efficient resource management \cite{wang-etal-2024-rethinking-bounds}, developing robust evaluation benchmarks \cite{liu2024agentbench, peng2024survey} and addressing ethical risks and safety concerns \cite{Akbulut2024, deshpande2023anthropomorphizationaiopportunitiesrisks, meinke2024frontiermodelscapableincontext}. 

\section{Method}

\begin{algorithm}[tb]
\caption{XpandA Workflow}
\label{alg:example}
\begin{algorithmic}[1] 
    \STATE {\bfseries Input:} context $C$, query $Q$
    \STATE {\bfseries Params:} LLM$(\theta)$, instruct prompts $(I_E, I_D)$, chunk size $K$, chunk overlap $P$, max replay times MRT
    \STATE Split $C$ into $x$ chunks $\{c_1, c_2, \ldots, c_x\}$
    \STATE Create empty dictionaries $T, P$
    \STATE Initialize $o \gets 1$, $rev \gets 1$, $\textit{replayCount} \gets 0$
    \REPEAT
        \FOR{$i$ in sequence $o, o+rev, o+2rev, \ldots$ (within chunk bounds $1..x$)} 
            \STATE $T_i, P_i \gets \text{Explorer}(I_E, Q, T, P, c_i)$
            \STATE $T \gets \text{Merge}(T, T_i)$
            \STATE $P \gets \text{Merge}(P, P_i)$
            \FOR{every key $k$ such that $k \in \text{keys}(P_i)$ and $k \in \text{keys}(T)$}
                \STATE Delete entry with key $k$ from $T$
            \ENDFOR
        \ENDFOR
        \STATE $\text{Action}, \hat{A} \gets \text{Decider}(I_D, Q, P)$
        \IF{$\text{"Replay"} \in \text{Action}$} 
            \STATE $o \gets \begin{cases}
                                \max( (\min \text{values}(T)) - 1, 1) & \text{if } rev > 0 \\
                                \min( (\max \text{values}(T)) + 1, x) & \text{otherwise} 
                            \end{cases}$
            \STATE $rev \gets -rev$
            \STATE $\textit{replayCount} \gets \textit{replayCount} + 1$
        \ELSE
            \STATE \textbf{Break}
        \ENDIF
    \UNTIL{$\textit{replayCount} >$ MRT}
    \STATE $A \gets \text{Parse}(\hat{A})$\;
    \STATE \textbf{return} $A$\;
\end{algorithmic}
\end{algorithm}

Figure~\ref{fig:overview} illustrates our XpandA framework, which operates in three stages. In Stage~1, Dynamic partition strategy is employed to decompose the input sequence of vastly varying length into chunks of appropriate length. In Stage~2, text chunks are assigned \textbf{one-to-one} to Explorers. Explorers then process the chunks sequentially by: decomposing problems, searching information and then generating new problems. Problems and relevant information are added to shared information memory. In Stage~3, Decider evaluates information completeness: if sufficient, it generates the final answer; otherwise, it proposes a selective replay trajectory of potentially informative chunks.

\subsection{Stage 1: Dynamic Partition of Long Input Sequence}
The Dynamic Partition method uses a saturation function: for shorter inputs, it increases unit chunk length; for longer inputs, the length saturates and chunk count grows. Let $w$ be the input token length, $n$ the target chunk count, $[L,K]$ the range of overlap, and $M$ the maximum chunk size. The partitioning follows:
\begin{equation}
\text{Chunk size} = 
\begin{cases}
\left\lceil \frac{w}{n} \right\rceil & \text{if } \frac{w}{n} \leq M \\
M & \text{if } \frac{w}{n} > M
\end{cases}
\end{equation}
where $\left\lceil \cdot \right\rceil$ denotes the ceiling function. In the first case, chunks maintain approximately equal size with an overlap $\delta$. In the second case, the number of chunks grows dynamically as:
\begin{equation}
\text{Number of chunks} = \left\lceil \frac{w}{M - \delta} \right\rceil
\end{equation}
where $\delta$ (with increase rate $\alpha$, upper limit $K$ and lower limit $L$) \footnote{The chunking strategy in Xpanda sets $n=3, L=10, K=2000, \alpha=0.1, M = 102400$.} equals to:
\begin{equation}
    \delta = max(L,min(\alpha*w, K))
\end{equation}
\subsection{Stage 2: Question-guided Exploration}
In Stage 1: XpandA operates on a sequence of $l$ Explorers. Each Explorer accepts the concatenation of a document chunk $c_i$, user query $q$, instruction for a specific reasoning task $I_D$ and gathered information, denoted as "(question,[answer]) pairs" $P$. Explorer is instructed to output enlarged $P_i$ and generate newly-generated problems included in "Unsolved Problem Tracer" $T_i$ expressed as:
\begin{equation}
    T_i, P_i \gets \text{Explorer}(I_E,Q,T,P,c_i)
\end{equation}
$T$,$P$ will be updated with $T_i$,$P_i$ when the processed chunk is switched. 
\begin{equation}
\begin{aligned}
    &T\gets Merge(T,T_i)\\
    &P\gets Merge(P,P_i)
\end{aligned}
\end{equation}
In the workflow, intricate problems are broken down into sub-problems $q_x^y$ (the $y^{th}$ problem from $x^{th}$ chunk), which are supplemented by new information, and then new problems are proposed based on the newly-acquired information $a_n^m$ (the $m^{th}$ answer for the $n^{th}$ problem). The process can be specified by:
\begin{equation}
\begin{aligned}
    & q_i^n \gets Expand(P_i)\\
    & a_n^j \gets Answer(q_x^n), \forall x<=i
\end{aligned}
\end{equation}
$Expand$ function has two types of behaviors: 1) Breaking down the generalized problem into sub-problems, and 2) expanding new problem entities based on a problem and the corresponding information. While the loop iterates, the problem-solving focus of large language model is gradually expanded across the chunks.
\subsection{Stage 3: Iterative Selective Replay}
Chunking inherently suffers from information loss: interrelated information across different chunks may be lost if not added to public memory, meaning the model cannot spontaneously notice dangling nodes in broken reasoning chains beyond its context window. For instance, if reasoning about information in chunk a depends on reasoning about information in chunk b (where $b<a$), the model cannot initiate reasoning for chunk b. To address this, we propose a heuristic replay strategy. The Unsolved Problem Tracer $T$ established in the previous stage is designed precisely for this strategy. Upon completion of each unidirectional reasoning phase, the Decider is activated to assess whether the currently gathered information suffices for reasoning and determines whether to initiate replay. 
\begin{equation}
   Action, \hat{A} \gets \text{Decider}(I_D,Q,P) 
\end{equation}The replay performs backward search from the last unsolved chunk, applying two heuristics: 1) the target chunk cannot answer all pending questions, and 2) all relevant information precedes it:
\begin{equation}
o \gets \begin{cases}
    \max(\min(T.values-1,1) & \text{if } rev>0 \\
    \min(\max(T.values+1,l) & \text{otherwise}
\end{cases}
\end{equation}
This dependency forms an AOV network where chunks are vertices and dependencies are edges. When MRT (max replay times) $> x-1$, XpandA guarantees to complete dependency resolution \ref{sec:proof} if LLM can finish information extraction in a chunk unit.

\section{Experiments}

\subsection{Experiment Setup}
\begin{table*}[t]
\caption{\textbf{Dataset Statistics}. Benchmark datasets from LongBench, $\infty$Bench, OpenAI.}
\resizebox{\textwidth}{!}{
\begin{tabular}{@{}lcccccccc@{}}
\toprule
\multicolumn{1}{l}{} & \multicolumn{2}{c}{Longbench} & \multicolumn{2}{c}{LV-Eval} & \multicolumn{2}{c}{$\infty$Bench} & \multicolumn{2}{c}{OpenAI}   \\ \cmidrule(lr){2-3}\cmidrule(lr){4-5}\cmidrule(lr){6-7}\cmidrule(lr){8-9}
     & HotpotQA &  NarrativeQA & Loogle-CR-mixup & Loogle-MR-mixup & Code.Run & Code.Debug & GraphWalks & OpenAI-MRCR  \\ \midrule
Avg len & 9151 & 18409  & 99.7k & 95.4k & 75.2k & 114.7k & 237.9k &  223.5k \\
Max len & 16k & 65k  & 256k & 256k & 80k & 200k & 1m & 1m\\
Task Type & Multi-doc QA & Single-doc QA  & Multi-hop QA & Multi-hop QA & Code & Code & Misc & Retrieval\\
Eval Metric & F1 & F1 &  keyword-recall-based F1 &  keyword-recall-based F1 & Accuracy & Accuracy & F1 & Seq Match \\ \bottomrule
\end{tabular}}
\vspace{-4mm}
\label{tab:dataset_stats}
\end{table*}

\begin{table*}[htbp]
\caption{{\bf Overall Performance evaluation} of XpandA on benchmarks for Question-Answering, Code, Miscellaneous, and Retrieval, showing XpandA achieved optimal or leading Performance over full-context, RAG and previous agent-based works across multiple domains. The darker blue areas indicate the globally optimal performance, while the lighter blue and red colors represent the local superiority and inferiority respectively.}
\label{tab:model_performance_comparison}
\centering
    
\medskip
\resizebox{\linewidth}{!}{
\begin{tabular}{@{}llcccccccc@{}}
\toprule
& & \multicolumn{4}{c}{\textbf{Question Answering}} & \multicolumn{2}{c}{\textbf{Code Tasks}} & \multicolumn{1}{c}{\textbf{Misc}} & \multicolumn{1}{c}{\textbf{Retrieval}} \\
\cmidrule(lr){3-6} \cmidrule(lr){7-8} \cmidrule(lr){9-9} \cmidrule(lr){10-10}
LLMs & Method & HotpotQA & Narr.QA & CR & MIR & Code.Debug & Code.Run & GraphWalks & MRCR \\
\midrule
{\textbf{Qwen2.5-7B-Inst}} & Full-context & 48.7 & 11.9 & 5.94 & 4.01 & 21.1 & 9.1 & 14.9 & 5.95 \\
& RAG & 53.8 & 14.2 & 7.54 & 5.29 & 22.1 & 3.2 & 25.6 & 8.8 \\
& XpandA & \cellcolor{blue!25}67.3 & \cellcolor{blue!25}20.7 & \cellcolor{blue!25}15.1 & \cellcolor{blue!25}12.6 & \cellcolor{blue!25} 24.8 & \cellcolor{blue!25}21.2 & \cellcolor{blue!25} 29.8 & \cellcolor{blue!25}16.3 \\
\midrule
{\textbf{LLama3.1-7B-Inst}} & Full-context & 53.2 & 12.3 & 8.3 & 7.41 & 20.2 & 7.7 & 14.9 & 2.09 \\
& RAG & 55.1 & 15.7 & \cellcolor{blue!15} 11.8 & 10.2 & \cellcolor{blue!15} 25.4 & 1.4 & \cellcolor{blue!15} 15.5 & 12.9 \\
& XpandA & \cellcolor{blue!25}69.4 & \cellcolor{blue!25}24.5 & \cellcolor{red!15} 10.1 & \cellcolor{blue!25}11.9 & \cellcolor{red!15} 24.3 & \cellcolor{blue!15} 20.6 & \cellcolor{red!15} 13.3 & \cellcolor{blue!25}14.5 \\
\midrule
{\textbf{Qwen2.5-7B-Inst-1M}} & Full-context & 56.5 & 20.1 & 8.78 & 7.87 & 27.8 & \cellcolor{blue!25}21.4 & 29.3 & 5.19 \\
& RAG & 57.1 & 20.3 & 9.42 & 7.49 & 25.3 & 2.3 & 27.6 & 5.57 \\
& CoA & 62.1 & \cellcolor{blue!25}25.3 & 13.4 & 9.7 & \cellcolor{blue!25}33.3 & 18.4 & \cellcolor{blue!25}30.3 &  11.9 \\
& LongAgent & 57.1 & 21.9 & \cellcolor{blue!25}14.2 & 8.8 & \cellcolor{blue!25}34.2 & 17.7 & \cellcolor{blue!25}31.1 & 9.85 \\
& XpandA & \cellcolor{blue!25}64.5 & \cellcolor{blue!25}26.4 & \cellcolor{blue!25}17.4 & \cellcolor{blue!25}13.9 & \cellcolor{blue!25}40.5 & \cellcolor{blue!25}25.8 & \cellcolor{blue!25}32.5 & \cellcolor{blue!25}15.3 \\
\bottomrule
\end{tabular}
}
\medskip
\end{table*}

\begin{figure*}
    \centering
    \includegraphics[width=0.95\linewidth]{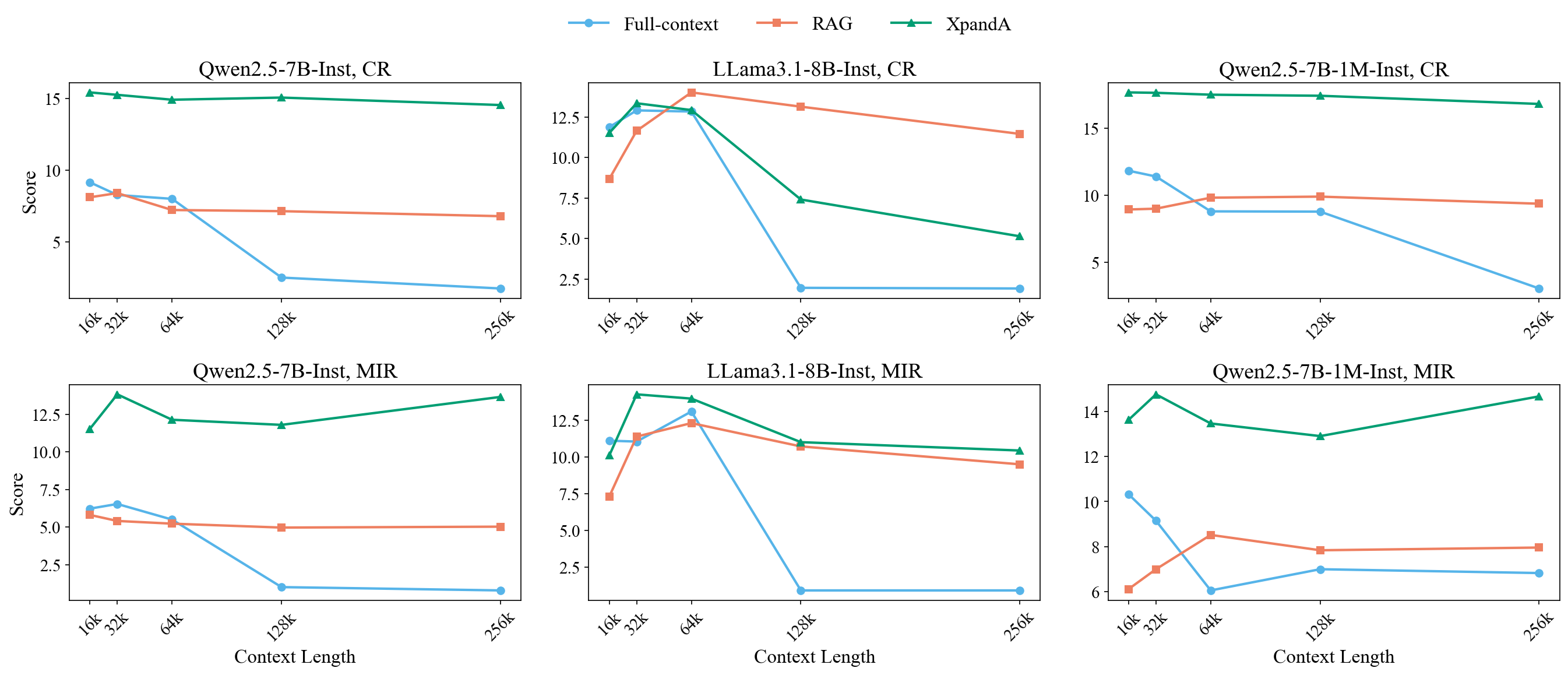}
    \caption{F1 Score of {\bf Full-context}, {\bf RAG} and {\bf XpandA} on all three backbone models in datasets \textbf{LooGLE-MIR-mixup} and \textbf{LooGLE-CR-mixup} of length varing from 16k to 256k. Xpanda achieves superiority on {\bf 5/6} datasets. Besides, Xpanda mitigates the performance decay in the case of input sequences of different lengths and reduces the gap between different generations of backbone models.}
    \label{fig:loogle}
\end{figure*}

\textbf{Datasets.} We conduct comprehensive experiments on long context datasets  from \textbf{LongBench} \cite{bai2024longbench}, \textbf{LV-Eval} \cite{yuan2024lveval}, $\infty$\textbf{Bench} \cite{zhang-etal-2024-bench} and two benchmarks \textbf{MRCR}\footnote{\href{https://huggingface.co/datasets/openai/mrcr}{https://huggingface.co/datasets/openai/mrcr}}, {\bf GraphWalks}\footnote{\href{https://huggingface.co/datasets/openai/graphwalks}{https://huggingface.co/datasets/openai/graphwalks}}  released by \textbf{OpenAI}  (Table~\ref{tab:dataset_stats}):
\begin{itemize}
    \item \textbf{Question Answering.} \textbf{HotpotQA} \cite{yang-etal-2018-hotpotqa} is a Wikipedia-based multihop QA dataset. It requires reasoning across multiple passages to find the answer. \textbf{NarrativeQA} \cite{narrativeqa} is a dataset focusing on answering questions based on stories or scripts, including understanding of important elements such as characters, plots, themes, etc. \textbf{loogle-MR-mixup} and \textbf{loogle-CR-mixup} originate from LooGLE \cite{li2024looglelongcontextlanguagemodels} Long-dependency QA task, specifically the Multiple information Retrieval and Comprehension and Reasoning subtasks and are mixed up with mix up distracting documents and supporting documents of stepwise growing length.
    \item \textbf{Retrieval.} \textbf{OpenAI-MRCR} \cite{vodrahalli2024michelangelolongcontextevaluations}is a long context dataset for benchmarking an LLM's ability to distinguish between multiple needles hidden in context. The evaluation consists of retrieval of the response corresponding to a specific instance from multi-turn synthetic dialogue. 
    \item \textbf{Code Tasks.} We consider two datasets from $\infty$Bench. \textbf{Code.Run} is a dataset that simulates multi-step function executions
 that involve basic arithmetic operations such as addition, subtraction, and nested function calls. \textbf{Code.Debug} is a multi-choice dataset to do the bug localization of recomposed repositories sourced from PyPI\footnote{\href{https://pypi.org/}{https://pypi.org/}}.
    \item \textbf{Misc.} We pick \textbf{GraphWalks}. Graphwalks is designed to require reasoning across multiple positions in the context and cannot be solved sequentially. In Graphwalks, the model is given a graph represented by its edge list and asked to perform an operation to search the parent or the breadth-first search (BFS) results from certain nodes.
\end{itemize}

\textbf{LLM Backbones.} The backbone models we deploy through include the well-established and popular open-source model Llama3.1-8B-Instuct \cite{grattafiori2024llama3herdmodels}, Qwen2.5-7B-Instruct \cite{qwen2025qwen25technicalreport} as well as the latest long-context language model Qwen2.5-7B-Instruct-1M \cite{yang2025qwen251mtechnicalreport} model. We choose efficient engine vllm \cite{kwon2023efficientmemorymanagementlarge} and 4*A100\_80G GPU as infrastructures for the experiments to handle immense GPU memory required for ultra-long input sequences up to 1M tokens.

\textbf{Baselines.} \textbf{Full-context} directly places the query and the long text into the context window where truncation may occur. \textbf{RAG} is implemented through RAG-Fusion \cite{Rackauckas_2024} by utilizing reciprocal rank fusion of relevance and word frequency ranking to reach strong and robust retrieval performance. We introduce previous \textbf{Agent-based} works CoA \cite{zhang2024chainagentslargelanguage} and LongAgent \cite{zhao2024longagentscalinglanguagemodels} to enhance the comprehensiveness of our baseline to highlight the improvement of our works. 

\textbf{Metrics.} We evaluate the QA tasks and Misc by {\bf F1} Score \cite{bai2024longbench} and Sequence Match Ratio\footnote{\href{https://docs.python.org/3/library/difflib.html}{https://docs.python.org/3/library/difflib.html}} for Retrieval, code tasks by {\bf exact match}. Besides, we introduce fine-grained {\bf Progress Score} $f$ from AgentBoard \cite{ma2024agentboard} for quantification of the  multi-turn LLM agents' performance which is denoted by:
\begin{equation}
    r_t = \max _{i, 0 \leq i \leq t}\frac{|\mathcal{G}\cap \mathcal{P}_i|}{|\mathcal{G}|}
\end{equation}
where an overall goal can be done by conjunction $g_1\land g_2, \dots, \land g_m$ of atomic subgoal $\mathcal{G}=\{g_1,g_2,\dots, g_m\}$. While $\mathcal{P}_i=\{p_1,p_2,\dots, p_m\}$ express processed subgoals at $i^{th}$ step.

\subsection{Overall Results of XpandA}
\textbf{Question Answering.} Table \ref{tab:model_performance_comparison} shows the results of QA tasks on three backbone models. XpandA outperformed the full-context models on all seven datasets, with improvements of up to 18.6\% on HotpotQA, 12.2\% on NarrativeQA, 9.16\% on LooGLE-CR-mixup, and 8.39\% on LooGLE-MIR-mixup, respectively. Furthermore, XpandA surpassed RAG systems that utilize the same backbone model as their generator in performance. Notably, XpandA's performance on QA tasks also exceeded that of previous agent-based methods, CoA and LongAgent, with gains of up to 5.1\%.

\textbf{Code Tasks, Retrieval \& Misc.} 
Table \ref{tab:model_performance_comparison} displays the test results for these three classification tasks. Similarly, XpandA achieved optimal scores on the evaluation metrics for the majority of these tasks. However, it is noteworthy that RAG + Llama3.1-8B-Instruct delivered the best performance on the GraphWalks dataset, with XpandA ranking second. This outcome may be attributed to the fact that the text in GraphWalks is presented in a structured "node->node" format (where each 'node' is a hexadecimal hash string). Such a structure potentially allows RAG to retrieve the intended textual information with greater precision compared to attention-based language models.

\subsection{Comparison with RAG and Long Context LLMs}  
Table \ref{tab:model_performance_comparison} shows that XpandA's optimal performance in eight datasets is greater than the optimal performance of RAG on the corresponding datasets, with increases of $3.4\%~22.6\%$ respectively. It is noted in Figure~\ref{fig:loogle} that Llama + RAG performs slightly better than Llama + XpandA, possibly due to Llama's weaker ability to follow the multi-agent workflow instructions in XpandA, leading to a performance decrease. Additionally, RAG performs poorly in multi-hop tasks, likely because it cannot iteratively retrieve and reason about information step-by-step, a core mechanism enabling XpandA to handle such problems effectively. For tasks involving 16k to 1M tokens, we evaluate Qwen2.5-7B-Instruct-1M, a long-context language model (LCLM) supporting up to 1 million tokens, against models with smaller context windows (SCLMs, typically 128k tokens). Figure~\ref{fig:loogle} shows that while LCLMs outperform SCLMs on 64k–256k token inputs, both exhibit performance degradation as sequences lengthen. XpandA mitigates this decay, maintaining stable F1 scores for 16k–256k inputs and enabling SCLMs to match LCLM performance on 256k-token tasks.

\subsection{Comparison with Other Multi-Agent Frameworks}
\begin{table}[t]
  \centering
  \caption{\textbf{Comparison with other agent-based works} on \textbf{GraphWalks} and \textbf{MRCR}. XpandA significantly outperforms LongAgent and CoA in both success (suc), progress rates (prog) and agent invoke steps (steps) on GraphWalks and MRCR, demonstrating superior efficiency in information acquisition and structured reasoning for multi-hop long-context problems.}
  \label{tab:agent_comparison_v2}
  \resizebox{\linewidth}{!}{
    \begin{tabular}{@{}llcccccc@{}}
    \toprule
    & & \multicolumn{2}{c}{\textbf{LongAgent}} & \multicolumn{2}{c}{\textbf{CoA}} & \multicolumn{2}{c}{\textbf{XpandA}} \\
    \cmidrule(lr){3-4} \cmidrule(lr){5-6} \cmidrule(lr){7-8}
    & & suc & prog/steps & suc & prog/steps & suc & prog/steps \\
    \midrule
    {\textbf{Graphwalks}} & Depth=2 & 53.2 & 63.2/6.4 & 55.6 & 71.8/3 & 62.4\textcolor{blue}{$_{\uparrow 6.8}$} & 80.2\textcolor{blue}{$_{\uparrow 8.4}$}/3.9 \\
    & Depth=4 & 21.4 & 27.7/8.5 & 6.25 & 21.4/3 & 38.2\textcolor{blue}{$_{\uparrow 16.8}$} & 53.6\textcolor{blue}{$_{\uparrow 25.9}$}/6.5 \\
    & Depth=8 & 6.7  & 10.2/8.9 & 2.3  & 11.4/3 & 19.4\textcolor{blue}{$_{\uparrow 12.7}$} & 42.9\textcolor{blue}{$_{\uparrow 31.5}$}/7.7 \\
    \midrule
    {\textbf{MRCR}} & 2 Needles & 61.3 & 63.7/6.1 & 66.4 & 68.6/3 & 71.8\textcolor{blue}{$_{\uparrow 5.4}$} & 77.8\textcolor{blue}{$_{\uparrow 9.2}$}/4.3 \\
    & 4 Needles & 25.7 & 25.4/8.2 & 29.4 & 54.3/3 & 39.1\textcolor{blue}{$_{\uparrow 9.7}$} & 68.3\textcolor{blue}{$_{\uparrow 14.0}$}/5.1 \\
    & 8 Needles & 5.1  & 12.3/8.8 & 19.4 & 40.5/3 & 20.6\textcolor{blue}{$_{\uparrow 1.2}$} & 44.8\textcolor{blue}{$_{\uparrow 4.3}$}/8.2 \\
    \bottomrule
    \end{tabular}%
  }
\end{table}
As shown in Table \ref{tab:agent_comparison_v2}, XpandA outperforms LongAgent and CoA in success rate by up to 16.8\% and progress rate by up to 32.7\% on GraphWalks, and by 13.4\% and 42.9\% respectively on MRCR. It is noteworthy that, when compared to CoA, which lacks a replay mechanism, XpandA's primary additional computational cost is the necessary overhead in operational steps due to its replay process. The higher progress rate reflects that XpandA more purposefully acquires information when traversing text chunks and more efficiently identifies the requisite information for each step in multi-hop problem-solving. The higher success rate signifies that XpandA develops a structured understanding of the collected information, thereby providing better answers rather than relying on superficial or decontextualized interpretations.

\section{Analysis}
\subsection{Diverse Chunking Strategies Meet Context of Vastly Varying Length}

\begin{figure}[h]
  \centering
  \includegraphics[width=\linewidth]{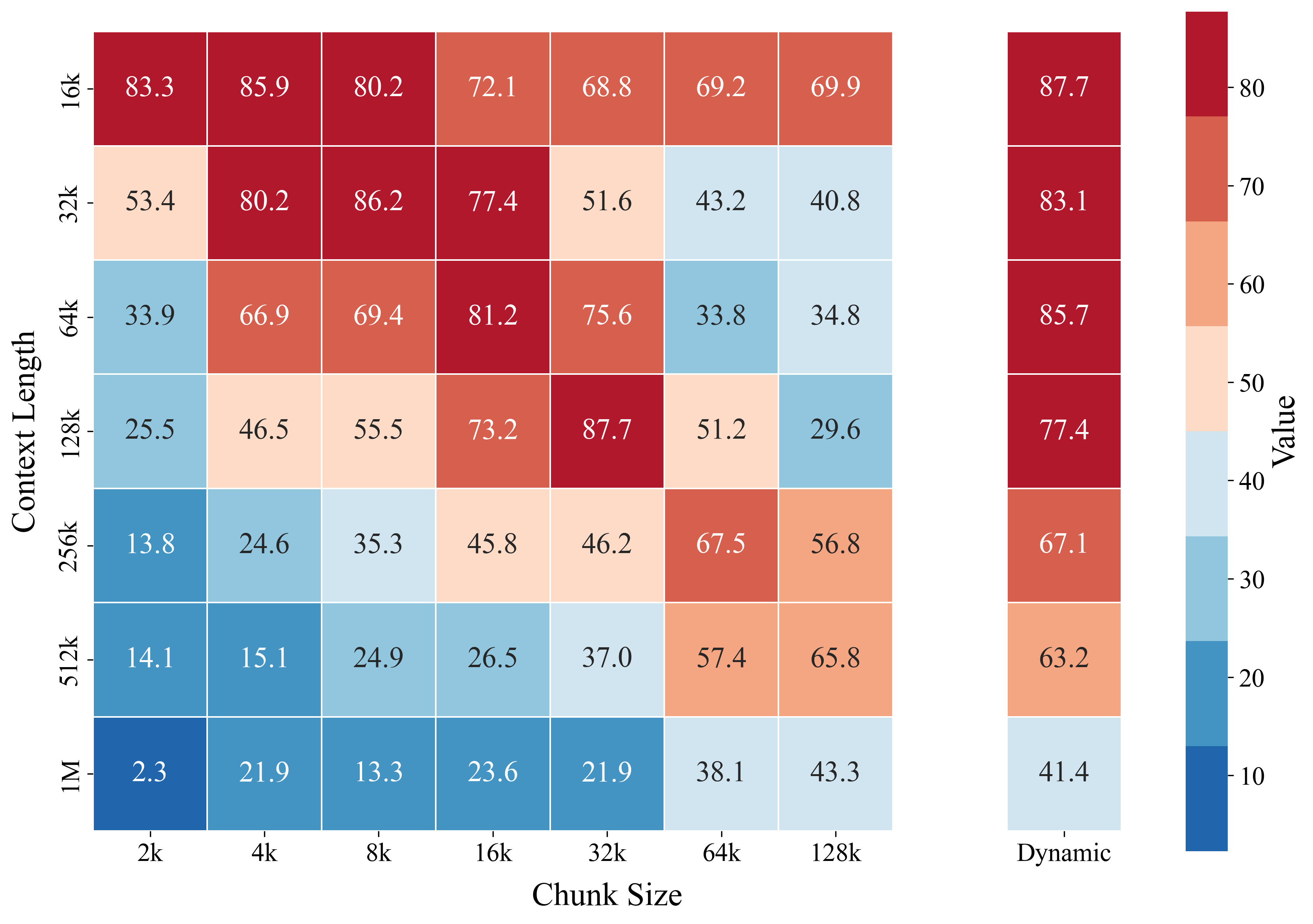} 
  \caption{Impact of different chunking strategies on Xpanda in dataset MRCR (2 Needle). Optimal static performance occurs when the unit chunk size is $[\frac{1}{16},\frac{1}{4}]$ of the input length, while dynamic chunking controls the proportion within this range and achieves 18\% average improvement.}
  \label{fig:heatmap}
\end{figure}

We analyze the impact of chunking strategies on XpandA's performance, shown in Figure~\ref{fig:heatmap}, via ablation experiments on the MRCR (2 Needles) benchmark (16k--1M sequences), evaluating retrieval success rate. Static chunking (2k--128k) shows that when the chunk size exceeds the input length, XpandA degenerates to a single-agent system, suffering full-context-like performance decay. Conversely, overly small chunks introduce excessive agents, collecting irrelevant noise and degrading accuracy. Optimal static performance occurs when unit chunk size is $[\frac{1}{16},\frac{1}{4}]$ of the input sequence length.\\
XpandA's dynamic chunking adaptively divides the input sequence into four to eight equal parts which fall into the range, thus achieving integrated peak performance of static strategy. This improves the average retrieval success rate by \textbf{18\%} over the best static variant.

\subsection{LLM Reasoning Scheme Ablation}


\begin{figure*}[t]
    \centering
    \begin{subfigure}[b]{0.3\textwidth}
        \centering
        \includegraphics[width=\textwidth]{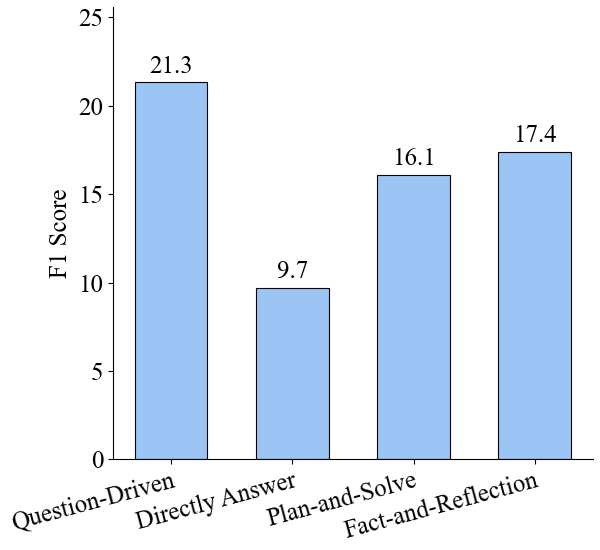}
        \caption{Reasoning scheme (F1 Score).}
        \label{fig:ablation_overall}
    \end{subfigure}
    \hfill
    \begin{subfigure}[b]{0.3\textwidth}
        \centering
        \includegraphics[width=\textwidth]{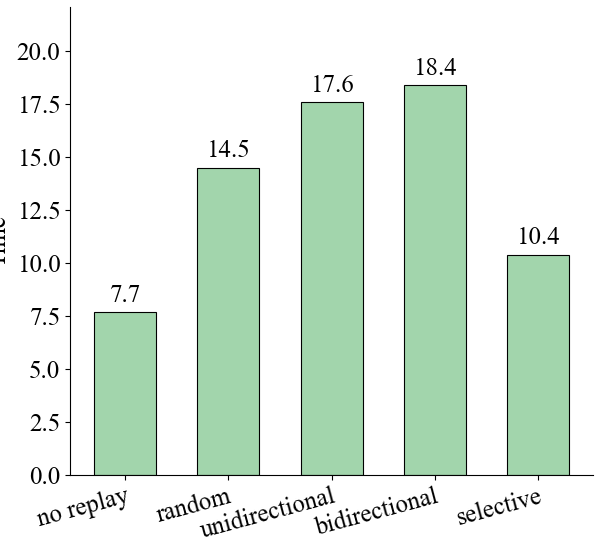}
        \caption{Replay strategy (Time).}
        \label{fig:ablation_replay_time}
    \end{subfigure}
    \hfill
    \begin{subfigure}[b]{0.3\textwidth}
        \centering
        \includegraphics[width=\textwidth]{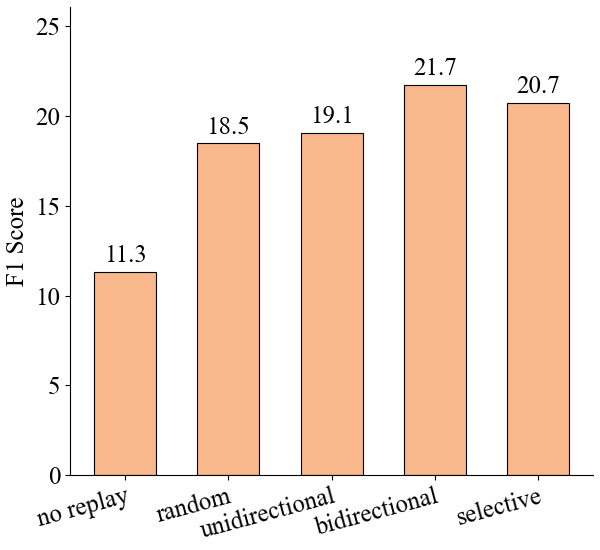}
        \caption{Replay strategy (F1 Score).}
        \label{fig:ablation_replay_f1}
    \end{subfigure}
    \caption{Ablation studies on NarrativeQA. (a) Comparison of XpandA with alternative planning schemes. (b) Impact of different replay strategies on average inference time. (c) Impact of different replay strategies on F1 score. Selective replay in XpandA achieves a strong balance between performance and efficiency.}
    \label{fig:ablation_studies_individual}
\end{figure*}
To evaluate XpandA's question-driven scheme, we modified components and observed performance, shown in Figure~\ref{fig:ablation_overall}. \textbf{Plan-and-Solve} \cite{wang2023planandsolvepromptingimprovingzeroshot} and \textbf{Fact-and-Reflection} \cite{zhao-etal-2024-fact}, similar to XpandA's subquestion state tracking, can replace its planning scheme. We also implemented \textbf{Directly Answer} by nullifying question trackers $P,T$ and having the Decider choose from Explorers' direct answers. Figure~\ref{fig:ablation_overall} shows XpandA outperforms these schemes by 4--12\% in F1 score on NarrativeQA. This indicates XpandA's question-driven scheme effectively enhances the multi-agent workflow by providing explicit instructions for context identification and exploration, mitigating inter-agent misalignment.

\subsection{Selective Replay Matters in Question-driven Workflow}


Figure~\ref{fig:ablation_replay_time}, \ref{fig:ablation_replay_f1} illustrate the performance of XpandA on the GraphWalks dataset (context length 128k) in terms of runtime and F1 score, comparing its standard selective replay mechanism against three alternative modes: no-replay, random replay, and brute-force replay (the latter executed either unidirectionally or by alternating directions with each replay). In the random replay setup, the number of chunks processed by selective replay was first recorded, and then an equivalent number of unique chunks was randomly selected and replayed sequentially. The experimental results indicate that selective replay achieves a nearly identical F1 score to that of the best-performing brute-force bidirectional replay, while utilizing only 56.5\% of the latter's runtime. The superiority in F1 score is primarily attributed to selective replay's fine-grained, atomic management of information states, which enables it to select chunks for replay that are most likely to uncover beneficial information and drive problem resolution. Furthermore, its enhanced time efficiency can also be ascribed to selective replay's ability to determine if the currently collected information is sufficient to answer the question, thereby enabling early stopping.

\subsection{Empirical Inference Time Analysis}
\begin{figure}[h] 
  \centering
  \includegraphics[width=0.9\linewidth]{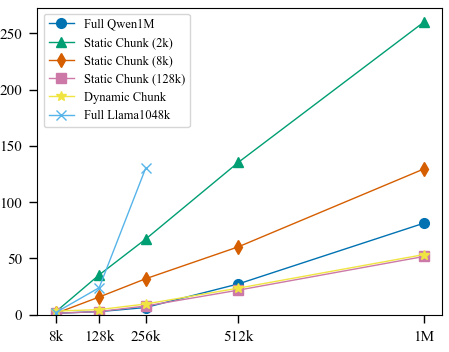} 
  \caption{XpandA scheme exhibits {\bf 1.5x speedup} in inference time compared to {\bf SOTA} long context language model on long texts. The x-axis represents the length of the context, and the y-axis represents the inference time (seconds).}
  \label{fig:time}
\end{figure}
We analyze the inference time of XpandA on 4×A100\_80G GPUs (Figure~\ref{fig:time}), using Qwen2.5:7B-Instruct-1M as the backbone to avoid API-related latency. For comparison, we include Llama3.1-Instruct-1048k \cite{gradientlongcontextllama3} in full-context mode. XpandA with dynamic chunking outperforms all baselines, achieving a 1.5× speedup over Qwen2.5-1M (with SOTA sparse attention optimization \cite{yang2025qwen251mtechnicalreport}) while maintaining linear complexity with fixed chunk sizes. In contrast, traditional full-context models (limited to 256k context on 4×A100) are significantly slower. Dynamic chunking also balances inference time and model-call latency better than static strategies (2k/8k) used in CoA \cite{zhang2024chainagentslargelanguage} and LongAgent \cite{zhao2024longagentscalinglanguagemodels}, which suffer from accumulated per-agent latency—especially critical for commercial API calls.

\section{Conclusion}
In this paper, we propose XpandA, a novel paradigm for multi-agent LLM workflow in long context language modeling through continuous questioning-answering. It is a plug-and-play, robust, interpretable, and efficient multi-agent system framework that introduces a dynamic chunking strategy and selective replay in its question-driven workflow to further enhance its performance. Extensive experiments show that XpandA achieves superior performance in various tasks with a wide span of context lengths (10k to 1M) compared to Full-context, RAG, and previous agent-based methods. It is worth noting that through XpandA expansion, SOTA (state-of-the-art) long context models have seen further improvements in performance (20\% metrics improvement and 1.5x speedup). Analysis indicates that XpandA, through the multi-module coupling of its question-driven workflow, effectively alleviates the failure problem of multi-agent systems, performing better in long-text processing scenarios.

\section{Impact Statement} 
This paper presents a framework for processing long context based on Multi-Agent System, without involving any social or ethical issues or consequences.

\section{Acknowledgements}
This work is supported by funding from the National Natural Science Foundation of China (NSFC No.62161160339).

\bibliographystyle{icml2025}
\bibliography{reference}

\begin{thebibliography}{110}
\providecommand{\natexlab}[1]{#1}
\providecommand{\url}[1]{\texttt{#1}}
\expandafter\ifx\csname urlstyle\endcsname\relax
  \providecommand{\doi}[1]{doi: #1}\else
  \providecommand{\doi}{doi: \begingroup \urlstyle{rm}\Url}\fi

\bibitem[Abdelnabi et~al.(2024)]{abdelnabi2024cooperation}
Abdelnabi, S. et~al.
\newblock Cooperation, competition, and maliciousness: {LLM}-stakeholders interactive negotiation.
\newblock In \emph{The Thirty-eight Conference on Neural Information Processing Systems Datasets and Benchmarks Track}, 2024.

\bibitem[Akbulut et~al.(2024)]{Akbulut2024}
Akbulut, C. et~al.
\newblock All too human? mapping and mitigating the risk from anthropomorphic ai.
\newblock \emph{Proceedings of the AAAI/ACM Conference on AI, Ethics, and Society}, 7:\penalty0 13–26, October 2024.
\newblock ISSN 3065-8365.

\bibitem[Aswani et~al.(2024)Aswani, Lu, Patankar, Dhalwani, Tan, Ganeshmohan, and Lacasse]{aswani-etal-2024-auto}
Aswani, K., Lu, H., Patankar, P., Dhalwani, P., Tan, X., Ganeshmohan, J., and Lacasse, S.
\newblock Auto-evolve: Enhancing large language model`s performance via self-reasoning framework.
\newblock In Al-Onaizan, Y., Bansal, M., and Chen, Y.-N. (eds.), \emph{Findings of the Association for Computational Linguistics: EMNLP 2024}, pp.\  13243--13257, Miami, Florida, USA, November 2024. Association for Computational Linguistics.
\newblock \doi{10.18653/v1/2024.findings-emnlp.774}.
\newblock URL \url{https://aclanthology.org/2024.findings-emnlp.774/}.

\bibitem[Bai et~al.(2024)Bai, Lv, Zhang, Lyu, Tang, Huang, Du, Liu, Zeng, Hou, Dong, Tang, and Li]{bai2024longbench}
Bai, Y., Lv, X., Zhang, J., Lyu, H., Tang, J., Huang, Z., Du, Z., Liu, X., Zeng, A., Hou, L., Dong, Y., Tang, J., and Li, J.
\newblock {L}ong{B}ench: A bilingual, multitask benchmark for long context understanding.
\newblock In \emph{Proceedings of the 62nd Annual Meeting of the Association for Computational Linguistics (Volume 1: Long Papers)}, pp.\  3119--3137, Bangkok, Thailand, August 2024. Association for Computational Linguistics.
\newblock \doi{10.18653/v1/2024.acl-long.172}.
\newblock URL \url{https://aclanthology.org/2024.acl-long.172}.

\bibitem[Beltagy et~al.(2020)Beltagy, Peters, and Cohan]{beltagy_longformer_2020}
Beltagy, I., Peters, M.~E., and Cohan, A.
\newblock Longformer: {The} {Long}-{Document} {Transformer}, December 2020.
\newblock URL \url{http://arxiv.org/abs/2004.05150}.
\newblock arXiv:2004.05150 [cs].

\bibitem[Besta et~al.(2024)Besta, Blach, Kubicek, Gerstenberger, Podstawski, Gianinazzi, Gajda, Lehmann, Niewiadomski, Nyczyk, et~al.]{besta2024graph}
Besta, M., Blach, N., Kubicek, A., Gerstenberger, R., Podstawski, M., Gianinazzi, L., Gajda, J., Lehmann, T., Niewiadomski, H., Nyczyk, P., et~al.
\newblock Graph of thoughts: Solving elaborate problems with large language models.
\newblock In \emph{Proceedings of the AAAI Conference on Artificial Intelligence}, volume~38, pp.\  17682--17690, 2024.

\bibitem[Bogomolov et~al.(2024)Bogomolov, Eliseeva, Galimzyanov, Glukhov, Shapkin, Tigina, Golubev, Kovrigin, van Deursen, Izadi, and Bryksin]{bogomolov2024longcodearenaset}
Bogomolov, E., Eliseeva, A., Galimzyanov, T., Glukhov, E., Shapkin, A., Tigina, M., Golubev, Y., Kovrigin, A., van Deursen, A., Izadi, M., and Bryksin, T.
\newblock Long code arena: a set of benchmarks for long-context code models, 2024.
\newblock URL \url{https://arxiv.org/abs/2406.11612}.

\bibitem[Borgeaud et~al.(2022)Borgeaud, Mensch, Hoffmann, Cai, Rutherford, Millican, van~den Driessche, Lespiau, Damoc, Clark, de~Las~Casas, Guy, Menick, Ring, Hennigan, Huang, Maggiore, Jones, Cassirer, Brock, Paganini, Irving, Vinyals, Osindero, Simonyan, Rae, Elsen, and Sifre]{borgeaud2022improvinglanguagemodelsretrieving}
Borgeaud, S., Mensch, A., Hoffmann, J., Cai, T., Rutherford, E., Millican, K., van~den Driessche, G., Lespiau, J.-B., Damoc, B., Clark, A., de~Las~Casas, D., Guy, A., Menick, J., Ring, R., Hennigan, T., Huang, S., Maggiore, L., Jones, C., Cassirer, A., Brock, A., Paganini, M., Irving, G., Vinyals, O., Osindero, S., Simonyan, K., Rae, J.~W., Elsen, E., and Sifre, L.
\newblock Improving language models by retrieving from trillions of tokens, 2022.
\newblock URL \url{https://arxiv.org/abs/2112.04426}.

\bibitem[Cemri et~al.(2025{\natexlab{a}})Cemri, Pan, Yang, Agrawal, Chopra, Tiwari, Keutzer, Parameswaran, Klein, Ramchandran, Zaharia, Gonzalez, and Stoica]{Cemri2025WhyDM}
Cemri, M., Pan, M.~Z., Yang, S., Agrawal, L.~A., Chopra, B., Tiwari, R., Keutzer, K., Parameswaran, A., Klein, D., Ramchandran, K., Zaharia, M., Gonzalez, J.~E., and Stoica, I.
\newblock Why do multi-agent llm systems fail?
\newblock \emph{ArXiv}, abs/2503.13657, 2025{\natexlab{a}}.
\newblock URL \url{https://api.semanticscholar.org/CorpusID:277103715}.

\bibitem[Cemri et~al.(2025{\natexlab{b}})Cemri, Pan, Yang, Agrawal, Chopra, Tiwari, Keutzer, Parameswaran, Klein, Ramchandran, Zaharia, Gonzalez, and Stoica]{cemri2025multiagentllmsystemsfail}
Cemri, M., Pan, M.~Z., Yang, S., Agrawal, L.~A., Chopra, B., Tiwari, R., Keutzer, K., Parameswaran, A., Klein, D., Ramchandran, K., Zaharia, M., Gonzalez, J.~E., and Stoica, I.
\newblock Why do multi-agent llm systems fail?, 2025{\natexlab{b}}.
\newblock URL \url{https://arxiv.org/abs/2503.13657}.

\bibitem[Chen et~al.(2023)Chen, Pasunuru, Weston, and Celikyilmaz]{chen2023walkingmemorymazecontext}
Chen, H., Pasunuru, R., Weston, J., and Celikyilmaz, A.
\newblock Walking down the memory maze: Beyond context limit through interactive reading, 2023.
\newblock URL \url{https://arxiv.org/abs/2310.05029}.

\bibitem[Chen et~al.(2024{\natexlab{a}})]{chen-etal-2024-llmarena}
Chen, J. et~al.
\newblock {LLMA}rena: Assessing capabilities of large language models in dynamic multi-agent environments.
\newblock In \emph{Proceedings of the 62nd Annual Meeting of the Association for Computational Linguistics (Volume 1: Long Papers)}, pp.\  13055--13077, Bangkok, Thailand, August 2024{\natexlab{a}}. Association for Computational Linguistics.

\bibitem[Chen et~al.(2024{\natexlab{b}})Chen, Zhang, and Han]{chen-etal-2024-comm}
Chen, P., Zhang, S., and Han, B.
\newblock {C}o{MM}: Collaborative multi-agent, multi-reasoning-path prompting for complex problem solving.
\newblock In Duh, K., Gomez, H., and Bethard, S. (eds.), \emph{Findings of the Association for Computational Linguistics: NAACL 2024}, pp.\  1720--1738, Mexico City, Mexico, June 2024{\natexlab{b}}. ACL.

\bibitem[Chen et~al.(2024{\natexlab{c}})]{chen2024agentverse}
Chen, W. et~al.
\newblock Agentverse: Facilitating multi-agent collaboration and exploring emergent behaviors.
\newblock In \emph{The Twelfth International Conference on Learning Representations}, 2024{\natexlab{c}}.

\bibitem[Dasigi et~al.(2021)Dasigi, Lo, Beltagy, Cohan, Smith, and Gardner]{Dasigi2021ADO}
Dasigi, P., Lo, K., Beltagy, I., Cohan, A., Smith, N.~A., and Gardner, M.
\newblock A dataset of information-seeking questions and answers anchored in research papers.
\newblock \emph{ArXiv}, abs/2105.03011, 2021.
\newblock URL \url{https://api.semanticscholar.org/CorpusID:234093776}.

\bibitem[Davidson et~al.(2024)]{davidson2024evaluating}
Davidson, T.~R. et~al.
\newblock Evaluating language model agency through negotiations.
\newblock In \emph{The Twelfth International Conference on Learning Representations}, 2024.

\bibitem[Deng et~al.(2024)Deng, Zhang, Oo, and Hooi]{deng-etal-2024-towards}
Deng, S., Zhang, N., Oo, N., and Hooi, B.
\newblock Towards a unified view of answer calibration for multi-step reasoning.
\newblock In Dalvi~Mishra, B., Durrett, G., Jansen, P., Lipkin, B., Neves~Ribeiro, D., Wong, L., Ye, X., and Zhao, W. (eds.), \emph{Proceedings of the 2nd Workshop on Natural Language Reasoning and Structured Explanations (@ACL 2024)}, pp.\  25--38, Bangkok, Thailand, August 2024. Association for Computational Linguistics.
\newblock URL \url{https://aclanthology.org/2024.nlrse-1.3/}.

\bibitem[Deshpande et~al.(2023)]{deshpande2023anthropomorphizationaiopportunitiesrisks}
Deshpande, A. et~al.
\newblock Anthropomorphization of ai: Opportunities and risks, 2023.

\bibitem[Dostál(2015)]{theoryps2015}
Dostál, J.
\newblock Theory of problem solving.
\newblock \emph{Procedia - Social and Behavioral Sciences}, 174:\penalty0 2798--2805, 2015.
\newblock ISSN 1877-0428.
\newblock \doi{https://doi.org/10.1016/j.sbspro.2015.01.970}.
\newblock URL \url{https://www.sciencedirect.com/science/article/pii/S1877042815010290}.
\newblock International Conference on New Horizons in Education, INTE 2014, 25-27 June 2014, Paris, France.

\bibitem[Dua et~al.(2022)Dua, Gupta, Singh, and Gardner]{dua-etal-2022-successive}
Dua, D., Gupta, S., Singh, S., and Gardner, M.
\newblock Successive prompting for decomposing complex questions.
\newblock In Goldberg, Y., Kozareva, Z., and Zhang, Y. (eds.), \emph{Proceedings of the 2022 Conference on Empirical Methods in Natural Language Processing}, pp.\  1251--1265, Abu Dhabi, United Arab Emirates, December 2022. Association for Computational Linguistics.
\newblock \doi{10.18653/v1/2022.emnlp-main.81}.
\newblock URL \url{https://aclanthology.org/2022.emnlp-main.81}.

\bibitem[Fan et~al.(2024)Fan, Sun, Xue, Zhang, Zhang, and Ruan]{fan_medodyssey_2024}
Fan, Y., Sun, H., Xue, K., Zhang, X., Zhang, S., and Ruan, T.
\newblock {MedOdyssey}: {A} {Medical} {Domain} {Benchmark} for {Long} {Context} {Evaluation} {Up} to {200K} {Tokens}, June 2024.
\newblock URL \url{http://arxiv.org/abs/2406.15019}.
\newblock arXiv:2406.15019 [cs].

\bibitem[Fu et~al.(2024{\natexlab{a}})Fu, Huang, Ning, Zhang, Chen, Wu, Wang, Huang, Li, Yan, Dai, Yang, and Wang]{fu2024moamixturesparseattention}
Fu, T., Huang, H., Ning, X., Zhang, G., Chen, B., Wu, T., Wang, H., Huang, Z., Li, S., Yan, S., Dai, G., Yang, H., and Wang, Y.
\newblock Moa: Mixture of sparse attention for automatic large language model compression, 2024{\natexlab{a}}.
\newblock URL \url{https://arxiv.org/abs/2406.14909}.

\bibitem[Fu et~al.(2024{\natexlab{b}})Fu, Panda, Niu, Yue, Hajishirzi, Kim, and Peng]{fu2024dataengineeringscalinglanguage}
Fu, Y., Panda, R., Niu, X., Yue, X., Hajishirzi, H., Kim, Y., and Peng, H.
\newblock Data engineering for scaling language models to 128k context, 2024{\natexlab{b}}.
\newblock URL \url{https://arxiv.org/abs/2402.10171}.

\bibitem[Gao et~al.(2025)Gao, Wu, Fu, and Hu]{gao2025questquerycentricdatasynthesis}
Gao, C., Wu, X., Fu, Q., and Hu, S.
\newblock Quest: Query-centric data synthesis approach for long-context scaling of large language model, 2025.
\newblock URL \url{https://arxiv.org/abs/2405.19846}.

\bibitem[Golovneva et~al.(2024)Golovneva, Wang, Weston, and Sukhbaatar]{Golovneva2024ContextualPE}
Golovneva, O., Wang, T., Weston, J.~E., and Sukhbaatar, S.
\newblock Contextual position encoding: Learning to count what's important.
\newblock \emph{ArXiv}, abs/2405.18719, 2024.
\newblock URL \url{https://api.semanticscholar.org/CorpusID:270094992}.

\bibitem[Grattafiori et~al.(2024)Grattafiori, Dubey, Jauhri, Pandey, Kadian, Al-Dahle, Letman, Mathur, Schelten, Vaughan, Yang, and Angela~Fan]{grattafiori2024llama3herdmodels}
Grattafiori, A., Dubey, A., Jauhri, A., Pandey, A., Kadian, A., Al-Dahle, A., Letman, A., Mathur, A., Schelten, A., Vaughan, A., Yang, A., and Angela~Fan, e.~a.
\newblock The llama 3 herd of models, 2024.
\newblock URL \url{https://arxiv.org/abs/2407.21783}.

\bibitem[He et~al.(2024)He, Pan, Dong, Song, LiuYiBo, Qianguosun, Liang, Wang, Zhang, and Zhang]{he-etal-2024-never}
He, J., Pan, K., Dong, X., Song, Z., LiuYiBo, L., Qianguosun, Q., Liang, Y., Wang, H., Zhang, E., and Zhang, J.
\newblock Never lost in the middle: Mastering long-context question answering with position-agnostic decompositional training.
\newblock In Ku, L.-W., Martins, A., and Srikumar, V. (eds.), \emph{Proceedings of the 62nd Annual Meeting of the Association for Computational Linguistics (Volume 1: Long Papers)}, pp.\  13628--13642, Bangkok, Thailand, August 2024. Association for Computational Linguistics.
\newblock \doi{10.18653/v1/2024.acl-long.736}.
\newblock URL \url{https://aclanthology.org/2024.acl-long.736/}.

\bibitem[Hong et~al.(2024)]{hong2024metagpt}
Hong, S. et~al.
\newblock Meta{GPT}: Meta programming for a multi-agent collaborative framework.
\newblock In \emph{The Twelfth International Conference on Learning Representations}, 2024.

\bibitem[Huang et~al.(2021)Huang, Cao, Parulian, Ji, and Wang]{huang-etal-2021-efficient}
Huang, L., Cao, S., Parulian, N., Ji, H., and Wang, L.
\newblock Efficient attentions for long document summarization.
\newblock In Toutanova, K., Rumshisky, A., Zettlemoyer, L., Hakkani-Tur, D., Beltagy, I., Bethard, S., Cotterell, R., Chakraborty, T., and Zhou, Y. (eds.), \emph{Proceedings of the 2021 Conference of the North American Chapter of the Association for Computational Linguistics: Human Language Technologies}, pp.\  1419--1436, Online, June 2021. Association for Computational Linguistics.
\newblock \doi{10.18653/v1/2021.naacl-main.112}.
\newblock URL \url{https://aclanthology.org/2021.naacl-main.112/}.

\bibitem[Jeyakumar et~al.(2024)Jeyakumar, Ahmad, and Gabriel]{jeyakumar2024advancing}
Jeyakumar, S.~K., Ahmad, A.~A., and Gabriel, A.~G.
\newblock Advancing agentic systems: Dynamic task decomposition, tool integration and evaluation using novel metrics and dataset.
\newblock In \emph{NeurIPS 2024 Workshop on Open-World Agents}, 2024.

\bibitem[Jiang et~al.(2023)Jiang, Ren, and Lin]{jiang-etal-2023-llm}
Jiang, D., Ren, X., and Lin, B.~Y.
\newblock {LLM}-blender: Ensembling large language models with pairwise ranking and generative fusion.
\newblock In \emph{Proceedings of the Annual Meeting of the Association for Computational Linguistics}, Jul. 2023.

\bibitem[Jimenez et~al.(2024)Jimenez, Yang, Wettig, Yao, Pei, Press, and Narasimhan]{jimenez2024swebench}
Jimenez, C.~E., Yang, J., Wettig, A., Yao, S., Pei, K., Press, O., and Narasimhan, K.~R.
\newblock {SWE}-bench: Can language models resolve real-world github issues?
\newblock In \emph{The Twelfth International Conference on Learning Representations}, 2024.
\newblock URL \url{https://openreview.net/forum?id=VTF8yNQM66}.

\bibitem[Jin et~al.(2023)Jin, Han, Yang, Jiang, Chang, and Hu]{jin2023growlengthacceleratingllmspretraining}
Jin, H., Han, X., Yang, J., Jiang, Z., Chang, C.-Y., and Hu, X.
\newblock Growlength: Accelerating llms pretraining by progressively growing training length, 2023.
\newblock URL \url{https://arxiv.org/abs/2310.00576}.

\bibitem[Joshi et~al.(2024)Joshi, Sarwar, Varshney, Nag, Agrawal, and Naik]{joshi2024reaperreasoningbasedretrieval}
Joshi, A., Sarwar, S.~M., Varshney, S., Nag, S., Agrawal, S., and Naik, J.
\newblock Reaper: Reasoning based retrieval planning for complex rag systems, 2024.
\newblock URL \url{https://arxiv.org/abs/2407.18553}.

\bibitem[Kocisk{\'y} et~al.(2017)Kocisk{\'y}, Schwarz, Blunsom, Dyer, Hermann, Melis, and Grefenstette]{Kocisk2017TheNR}
Kocisk{\'y}, T., Schwarz, J., Blunsom, P., Dyer, C., Hermann, K.~M., Melis, G., and Grefenstette, E.
\newblock The narrativeqa reading comprehension challenge.
\newblock \emph{Transactions of the Association for Computational Linguistics}, 6:\penalty0 317--328, 2017.
\newblock URL \url{https://api.semanticscholar.org/CorpusID:2593903}.

\bibitem[Kojima et~al.(2022)Kojima, Gu, Reid, Matsuo, and Iwasawa]{kojima2022large}
Kojima, T., Gu, S.~S., Reid, M., Matsuo, Y., and Iwasawa, Y.
\newblock Large language models are zero-shot reasoners.
\newblock \emph{Advances in neural information processing systems}, 35:\penalty0 22199--22213, 2022.

\bibitem[Kwon et~al.(2023)Kwon, Li, Zhuang, Sheng, Zheng, Yu, Gonzalez, Zhang, and Stoica]{kwon2023efficientmemorymanagementlarge}
Kwon, W., Li, Z., Zhuang, S., Sheng, Y., Zheng, L., Yu, C.~H., Gonzalez, J.~E., Zhang, H., and Stoica, I.
\newblock Efficient memory management for large language model serving with pagedattention, 2023.
\newblock URL \url{https://arxiv.org/abs/2309.06180}.

\bibitem[Lake et~al.(2016)Lake, Ullman, Tenenbaum, and Gershman]{lake2016buildingmachineslearnthink}
Lake, B.~M., Ullman, T.~D., Tenenbaum, J.~B., and Gershman, S.~J.
\newblock Building machines that learn and think like people, 2016.
\newblock URL \url{https://arxiv.org/abs/1604.00289}.

\bibitem[Lee et~al.(2024)Lee, Chen, Dai, Dua, Sachan, Boratko, Luan, Arnold, Perot, Dalmia, Hu, Lin, Pasupat, Amini, Cole, Riedel, Naim, Chang, and Guu]{Lee2024CanLL}
Lee, J., Chen, A., Dai, Z., Dua, D., Sachan, D.~S., Boratko, M., Luan, Y., Arnold, S. M.~R., Perot, V., Dalmia, S., Hu, H., Lin, X., Pasupat, P., Amini, A., Cole, J.~R., Riedel, S., Naim, I., Chang, M.-W., and Guu, K.
\newblock Can long-context language models subsume retrieval, rag, sql, and more?
\newblock \emph{ArXiv}, abs/2406.13121, 2024.
\newblock URL \url{https://api.semanticscholar.org/CorpusID:270620072}.

\bibitem[Li et~al.(2024{\natexlab{a}})]{li2024cultureparkboostingcrossculturalunderstanding}
Li, C. et~al.
\newblock Culturepark: Boosting cross-cultural understanding in large language models, 2024{\natexlab{a}}.

\bibitem[Li et~al.(2023{\natexlab{a}})]{li2023camel}
Li, G. et~al.
\newblock {CAMEL}: Communicative agents for ''mind'' exploration of large language model society.
\newblock In \emph{Thirty-seventh Conference on Neural Information Processing Systems}, 2023{\natexlab{a}}.

\bibitem[Li et~al.(2023{\natexlab{b}})]{li-etal-2023-theory}
Li, H. et~al.
\newblock Theory of mind for multi-agent collaboration via large language models.
\newblock In Bouamor, H., Pino, J., and Bali, K. (eds.), \emph{Proceedings of the 2023 Conference on Empirical Methods in Natural Language Processing}, pp.\  180--192, Singapore, December 2023{\natexlab{b}}. Association for Computational Linguistics.

\bibitem[Li et~al.(2024{\natexlab{b}})Li, Wang, Zheng, and Zhang]{li2024looglelongcontextlanguagemodels}
Li, J., Wang, M., Zheng, Z., and Zhang, M.
\newblock Loogle: Can long-context language models understand long contexts?, 2024{\natexlab{b}}.
\newblock URL \url{https://arxiv.org/abs/2311.04939}.

\bibitem[Li et~al.(2024{\natexlab{c}})Li, Zhang, Yu, Fu, and Ye]{li2024more}
Li, J., Zhang, Q., Yu, Y., Fu, Q., and Ye, D.
\newblock More agents is all you need.
\newblock \emph{Transactions on Machine Learning Research}, 2024{\natexlab{c}}.
\newblock ISSN 2835-8856.
\newblock URL \url{https://openreview.net/forum?id=bgzUSZ8aeg}.

\bibitem[Li et~al.(2024{\natexlab{d}})Li, He, Guo, Bu, Bai, Liu, Liu, Qu, Li, Ouyang, Su, and Zheng]{li2024graphreaderbuildinggraphbasedagent}
Li, S., He, Y., Guo, H., Bu, X., Bai, G., Liu, J., Liu, J., Qu, X., Li, Y., Ouyang, W., Su, W., and Zheng, B.
\newblock Graphreader: Building graph-based agent to enhance long-context abilities of large language models, 2024{\natexlab{d}}.
\newblock URL \url{https://arxiv.org/abs/2406.14550}.

\bibitem[Li et~al.(2024{\natexlab{e}})Li, Zhang, Do, Yue, and Chen]{Li2024LongcontextLS}
Li, T., Zhang, G., Do, Q.~D., Yue, X., and Chen, W.
\newblock Long-context llms struggle with long in-context learning.
\newblock \emph{ArXiv}, abs/2404.02060, 2024{\natexlab{e}}.
\newblock URL \url{https://api.semanticscholar.org/CorpusID:268857023}.

\bibitem[Lieber et~al.(2024)Lieber, Lenz, Bata, Cohen, Osin, Dalmedigos, Safahi, Meirom, Belinkov, Shalev-Shwartz, Abend, Alon, Asida, Bergman, Glozman, Gokhman, Manevich, Ratner, Rozen, Shwartz, Zusman, and Shoham]{lieber2024jambahybridtransformermambalanguage}
Lieber, O., Lenz, B., Bata, H., Cohen, G., Osin, J., Dalmedigos, I., Safahi, E., Meirom, S., Belinkov, Y., Shalev-Shwartz, S., Abend, O., Alon, R., Asida, T., Bergman, A., Glozman, R., Gokhman, M., Manevich, A., Ratner, N., Rozen, N., Shwartz, E., Zusman, M., and Shoham, Y.
\newblock Jamba: A hybrid transformer-mamba language model, 2024.
\newblock URL \url{https://arxiv.org/abs/2403.19887}.

\bibitem[Liu et~al.(2025{\natexlab{a}})Liu, Zhu, Bai, He, Liao, Que, Wang, Zhang, Zhang, Zhang, Zhang, Chen, Guo, Li, Liu, Shan, Song, Tian, Wu, Zhou, Zhu, Feng, Gao, He, Li, Liu, Meng, Su, Tan, Wang, Yang, Ye, Zheng, Zhou, Huang, Li, and Zhang]{liu2025comprehensivesurveylongcontext}
Liu, J., Zhu, D., Bai, Z., He, Y., Liao, H., Que, H., Wang, Z., Zhang, C., Zhang, G., Zhang, J., Zhang, Y., Chen, Z., Guo, H., Li, S., Liu, Z., Shan, Y., Song, Y., Tian, J., Wu, W., Zhou, Z., Zhu, R., Feng, J., Gao, Y., He, S., Li, Z., Liu, T., Meng, F., Su, W., Tan, Y., Wang, Z., Yang, J., Ye, W., Zheng, B., Zhou, W., Huang, W., Li, S., and Zhang, Z.
\newblock A comprehensive survey on long context language modeling, 2025{\natexlab{a}}.
\newblock URL \url{https://arxiv.org/abs/2503.17407}.

\bibitem[Liu et~al.(2025{\natexlab{b}})Liu, Zhu, Bai, He, Liao, Que, Wang, Zhang, Zhang, Zhang, Zhang, Chen, Guo, Li, Liu, Shan, Song, Tian, Wu, Zhou, Zhu, Feng, Gao, He, Li, Liu, Meng, Su, Tan, Wang, Yang, Ye, Zheng, Zhou, Huang, Li, and Zhang]{Liu2025ACS}
Liu, J., Zhu, D., Bai, Z., He, Y., Liao, H., Que, H., Wang, Z.~M., Zhang, C., Zhang, G., Zhang, J., Zhang, Y., Chen, Z., Guo, H., Li, S., Liu, Z., Shan, Y., Song, Y., Tian, J., Wu, W., Zhou, Z., Zhu, R., Feng, J., Gao, Y., He, S., Li, Z., Liu, T., Meng, F., Su, W., Tan, Y., Wang, Z., Yang, J., Ye, W., Zheng, B., Zhou, W., Huang, W., Li, S., and Zhang, Z.
\newblock A comprehensive survey on long context language modeling.
\newblock \emph{ArXiv}, abs/2503.17407, 2025{\natexlab{b}}.
\newblock URL \url{https://api.semanticscholar.org/CorpusID:277271533}.

\bibitem[Liu et~al.(2023)Liu, Lin, Hewitt, Paranjape, Bevilacqua, Petroni, and Liang]{liu2023lostmiddlelanguagemodels}
Liu, N.~F., Lin, K., Hewitt, J., Paranjape, A., Bevilacqua, M., Petroni, F., and Liang, P.
\newblock Lost in the middle: How language models use long contexts, 2023.
\newblock URL \url{https://arxiv.org/abs/2307.03172}.

\bibitem[Liu et~al.(2024{\natexlab{a}})]{liu2024agentbench}
Liu, X. et~al.
\newblock Agentbench: Evaluating {LLM}s as agents.
\newblock In \emph{The Twelfth International Conference on Learning Representations}, 2024{\natexlab{a}}.

\bibitem[Liu et~al.(2024{\natexlab{b}})]{liu2024a}
Liu, Z. et~al.
\newblock A dynamic {LLM}-powered agent network for task-oriented agent collaboration.
\newblock In \emph{First Conference on Language Modeling}, Oct. 2024{\natexlab{b}}.

\bibitem[Lu et~al.(2025)Lu, Jiang, Liu, Du, Jiang, Hong, Liu, He, Yuan, Wang, Huang, Yuan, Xu, Xu, Lai, Chen, Zheng, Yan, Su, Wu, Zhang, Yang, Zhou, Zhang, and Qiu]{lu2025mobamixtureblockattention}
Lu, E., Jiang, Z., Liu, J., Du, Y., Jiang, T., Hong, C., Liu, S., He, W., Yuan, E., Wang, Y., Huang, Z., Yuan, H., Xu, S., Xu, X., Lai, G., Chen, Y., Zheng, H., Yan, J., Su, J., Wu, Y., Zhang, N.~Y., Yang, Z., Zhou, X., Zhang, M., and Qiu, J.
\newblock Moba: Mixture of block attention for long-context llms, 2025.
\newblock URL \url{https://arxiv.org/abs/2502.13189}.

\bibitem[Ma et~al.(2024)Ma, Zhang, Zhu, Yang, Yang, Jin, Lan, Kong, and He]{ma2024agentboard}
Ma, C., Zhang, J., Zhu, Z., Yang, C., Yang, Y., Jin, Y., Lan, Z., Kong, L., and He, J.
\newblock Agentboard: An analytical evaluation board of multi-turn llm agents, 2024.

\bibitem[Masry \& Hajian(2024)Masry and Hajian]{masry_longfin_2024}
Masry, A. and Hajian, A.
\newblock {LongFin}: {A} {Multimodal} {Document} {Understanding} {Model} for {Long} {Financial} {Domain} {Documents}, January 2024.
\newblock URL \url{http://arxiv.org/abs/2401.15050}.
\newblock arXiv:2401.15050 [cs].

\bibitem[Mehta et~al.(2023)Mehta, Gupta, Tay, Dehghani, Tran, Rao, Najork, Strubell, and Metzler]{mehta2023dsiupdatingtransformermemory}
Mehta, S.~V., Gupta, J., Tay, Y., Dehghani, M., Tran, V.~Q., Rao, J., Najork, M., Strubell, E., and Metzler, D.
\newblock Dsi++: Updating transformer memory with new documents, 2023.
\newblock URL \url{https://arxiv.org/abs/2212.09744}.

\bibitem[Meinke et~al.(2024)]{meinke2024frontiermodelscapableincontext}
Meinke, A. et~al.
\newblock Frontier models are capable of in-context scheming, 2024.

\bibitem[Morris et~al.(2024)Morris, Sohl-dickstein, Fiedel, Warkentin, Dafoe, Faust, Farabet, and Legg]{morris2024levelsagioperationalizingprogress}
Morris, M.~R., Sohl-dickstein, J., Fiedel, N., Warkentin, T., Dafoe, A., Faust, A., Farabet, C., and Legg, S.
\newblock Levels of agi for operationalizing progress on the path to agi, 2024.
\newblock URL \url{https://arxiv.org/abs/2311.02462}.

\bibitem[Munkhdalai et~al.(2024)Munkhdalai, Faruqui, and Gopal]{munkhdalai_leave_2024}
Munkhdalai, T., Faruqui, M., and Gopal, S.
\newblock Leave {No} {Context} {Behind}: {Efficient} {Infinite} {Context} {Transformers} with {Infini}-attention, August 2024.
\newblock URL \url{http://arxiv.org/abs/2404.07143}.
\newblock arXiv:2404.07143 [cs].

\bibitem[Nguyen et~al.(2024)Nguyen, Razniewski, and Weikum]{10.1145/3627673.3679768}
Nguyen, T.-P., Razniewski, S., and Weikum, G.
\newblock Cultural commonsense knowledge for intercultural dialogues.
\newblock In \emph{Proceedings of the 33rd ACM International Conference on Information and Knowledge Management}, CIKM '24, pp.\  1774–1784, New York, NY, USA, 2024. Association for Computing Machinery.
\newblock ISBN 9798400704369.

\bibitem[Ning et~al.(2024)]{ning2024skeletonofthought}
Ning, X. et~al.
\newblock Skeleton-of-thought: Prompting {LLM}s for efficient parallel generation.
\newblock In \emph{The Twelfth International Conference on Learning Representations}, 2024.

\bibitem[Pekelis et~al.(2024)Pekelis, Feil, Moret, Huang, and Peng]{gradientlongcontextllama3}
Pekelis, L., Feil, M., Moret, F., Huang, M., and Peng, T.
\newblock Llama 3 gradient: A series of long context models, 2024.
\newblock URL \url{https://gradient.ai/blog/scaling-rotational-embeddings-for-long-context-language-models}.

\bibitem[Peng et~al.(2024)]{peng2024survey}
Peng, J.-L. et~al.
\newblock A survey of useful llm evaluation.
\newblock \emph{arXiv preprint arXiv:2406.00936}, 2024.

\bibitem[Press et~al.(2021)Press, Smith, and Lewis]{Press2021TrainST}
Press, O., Smith, N.~A., and Lewis, M.
\newblock Train short, test long: Attention with linear biases enables input length extrapolation.
\newblock \emph{ArXiv}, abs/2108.12409, 2021.
\newblock URL \url{https://api.semanticscholar.org/CorpusID:237347130}.

\bibitem[Qin et~al.(2024)Qin, Sun, Li, Shen, Sun, and Zhong]{qin_lightning_2024}
Qin, Z., Sun, W., Li, D., Shen, X., Sun, W., and Zhong, Y.
\newblock Lightning {Attention}-2: {A} {Free} {Lunch} for {Handling} {Unlimited} {Sequence} {Lengths} in {Large} {Language} {Models}, January 2024.
\newblock URL \url{http://arxiv.org/abs/2401.04658}.
\newblock arXiv:2401.04658 [cs].

\bibitem[Qwen et~al.(2025)Qwen, :, Yang, Yang, Zhang, Hui, Zheng, Yu, Li, Liu, Huang, Wei, Lin, Yang, Tu, Zhang, Yang, Yang, Zhou, Lin, Dang, Lu, Bao, Yang, Yu, Li, Xue, Zhang, Zhu, Men, Lin, Li, Tang, Xia, Ren, Ren, Fan, Su, Zhang, Wan, Liu, Cui, Zhang, and Qiu]{qwen2025qwen25technicalreport}
Qwen, :, Yang, A., Yang, B., Zhang, B., Hui, B., Zheng, B., Yu, B., Li, C., Liu, D., Huang, F., Wei, H., Lin, H., Yang, J., Tu, J., Zhang, J., Yang, J., Yang, J., Zhou, J., Lin, J., Dang, K., Lu, K., Bao, K., Yang, K., Yu, L., Li, M., Xue, M., Zhang, P., Zhu, Q., Men, R., Lin, R., Li, T., Tang, T., Xia, T., Ren, X., Ren, X., Fan, Y., Su, Y., Zhang, Y., Wan, Y., Liu, Y., Cui, Z., Zhang, Z., and Qiu, Z.
\newblock Qwen2.5 technical report, 2025.
\newblock URL \url{https://arxiv.org/abs/2412.15115}.

\bibitem[Rackauckas(2024)]{Rackauckas_2024}
Rackauckas, Z.
\newblock Rag-fusion: A new take on retrieval augmented generation.
\newblock \emph{International Journal on Natural Language Computing}, 13\penalty0 (1):\penalty0 37–47, February 2024.
\newblock ISSN 2319-4111.
\newblock \doi{10.5121/ijnlc.2024.13103}.
\newblock URL \url{http://dx.doi.org/10.5121/ijnlc.2024.13103}.

\bibitem[s~Ko\v~cisk\'y et~al.(2018)s~Ko\v~cisk\'y, Schwarz, Blunsom, Dyer, Hermann, Melis, and Grefenstette]{narrativeqa}
s~Ko\v~cisk\'y, T., Schwarz, J., Blunsom, P., Dyer, C., Hermann, K.~M., Melis, G., and Grefenstette, E.
\newblock The {NarrativeQA} reading comprehension challenge.
\newblock \emph{Transactions of the Association for Computational Linguistics}, TBD:\penalty0 TBD, 2018.
\newblock URL \url{https://TBD}.

\bibitem[Shayegani et~al.(2023)]{shayegani2023survey}
Shayegani, E. et~al.
\newblock Survey of vulnerabilities in large language models revealed by adversarial attacks.
\newblock \emph{arXiv preprint arXiv:2310.10844}, 2023.

\bibitem[Shinn et~al.(2023)]{shinn2023reflexion}
Shinn, N. et~al.
\newblock Reflexion: language agents with verbal reinforcement learning.
\newblock In \emph{Thirty-seventh Conference on Neural Information Processing Systems}, 2023.

\bibitem[Si et~al.(2025)Si, Zhao, Chen, Li, Luo, Lv, An, Qi, Chang, and Sun]{si_gateau_2025}
Si, S., Zhao, H., Chen, G., Li, Y., Luo, K., Lv, C., An, K., Qi, F., Chang, B., and Sun, M.
\newblock {GATEAU}: {Selecting} {Influential} {Samples} for {Long} {Context} {Alignment}, February 2025.
\newblock URL \url{http://arxiv.org/abs/2410.15633}.
\newblock arXiv:2410.15633 [cs].

\bibitem[Stanovich \& West(2000)Stanovich and West]{stanovich2000individual}
Stanovich, K.~E. and West, R.~F.
\newblock Individual differences in reasoning: {I}mplications for the rationality debate?
\newblock \emph{Behavioral and Brain Sciences}, 23\penalty0 (5):\penalty0 645--665, 2000.

\bibitem[Su et~al.(2021)Su, Lu, Pan, Wen, and Liu]{Su2021RoFormerET}
Su, J., Lu, Y., Pan, S., Wen, B., and Liu, Y.
\newblock Roformer: Enhanced transformer with rotary position embedding.
\newblock \emph{ArXiv}, abs/2104.09864, 2021.
\newblock URL \url{https://api.semanticscholar.org/CorpusID:233307138}.

\bibitem[Sun et~al.(2022)Sun, Dong, Patra, Ma, Huang, Benhaim, Chaudhary, Song, and Wei]{Sun2022ALT}
Sun, Y., Dong, L., Patra, B., Ma, S., Huang, S., Benhaim, A., Chaudhary, V., Song, X., and Wei, F.
\newblock A length-extrapolatable transformer.
\newblock \emph{ArXiv}, abs/2212.10554, 2022.
\newblock URL \url{https://api.semanticscholar.org/CorpusID:254877252}.

\bibitem[Suzgun \& Kalai(2024)Suzgun and Kalai]{suzgun2024metapromptingenhancinglanguagemodels}
Suzgun, M. and Kalai, A.~T.
\newblock Meta-prompting: Enhancing language models with task-agnostic scaffolding, 2024.

\bibitem[Vodrahalli et~al.(2024)Vodrahalli, Ontanon, Tripuraneni, Xu, Jain, Shivanna, Hui, Dikkala, Kazemi, Fatemi, Anil, Dyer, Shakeri, Vij, Mehta, Ramasesh, Le, Chi, Lu, Firat, Lazaridou, Lespiau, Attaluri, and Olszewska]{vodrahalli2024michelangelolongcontextevaluations}
Vodrahalli, K., Ontanon, S., Tripuraneni, N., Xu, K., Jain, S., Shivanna, R., Hui, J., Dikkala, N., Kazemi, M., Fatemi, B., Anil, R., Dyer, E., Shakeri, S., Vij, R., Mehta, H., Ramasesh, V., Le, Q., Chi, E., Lu, Y., Firat, O., Lazaridou, A., Lespiau, J.-B., Attaluri, N., and Olszewska, K.
\newblock Michelangelo: Long context evaluations beyond haystacks via latent structure queries, 2024.
\newblock URL \url{https://arxiv.org/abs/2409.12640}.

\bibitem[Wang et~al.(2025)Wang, Ning, Pan, Wu, Guo, Deng, Bao, Hu, Zhang, Wang, and Zhang]{wang2025novelqabenchmarkingquestionanswering}
Wang, C., Ning, R., Pan, B., Wu, T., Guo, Q., Deng, C., Bao, G., Hu, X., Zhang, Z., Wang, Q., and Zhang, Y.
\newblock Novelqa: Benchmarking question answering on documents exceeding 200k tokens, 2025.
\newblock URL \url{https://arxiv.org/abs/2403.12766}.

\bibitem[Wang et~al.(2024{\natexlab{a}})Wang, Prasad, Stengel-Eskin, and Bansal]{wang2024soft}
Wang, H., Prasad, A., Stengel-Eskin, E., and Bansal, M.
\newblock Soft self-consistency improves language model agents.
\newblock \emph{arXiv preprint arXiv:2402.13212}, 2024{\natexlab{a}}.

\bibitem[Wang et~al.(2023{\natexlab{a}})Wang, Xu, Lan, Hu, Lan, Lee, and Lim]{wang2023planandsolvepromptingimprovingzeroshot}
Wang, L., Xu, W., Lan, Y., Hu, Z., Lan, Y., Lee, R. K.-W., and Lim, E.-P.
\newblock Plan-and-solve prompting: Improving zero-shot chain-of-thought reasoning by large language models, 2023{\natexlab{a}}.
\newblock URL \url{https://arxiv.org/abs/2305.04091}.

\bibitem[Wang et~al.(2024{\natexlab{b}})]{wang-etal-2024-rethinking-bounds}
Wang, Q. et~al.
\newblock Rethinking the bounds of {LLM} reasoning: Are multi-agent discussions the key?
\newblock In \emph{Proceedings of the Annual Meeting of the Association for Computational Linguistics}, Aug. 2024{\natexlab{b}}.

\bibitem[Wang et~al.(2023{\natexlab{b}})Wang, Dong, Cheng, Liu, Yan, Gao, and Wei]{wang2023augmentinglanguagemodelslongterm}
Wang, W., Dong, L., Cheng, H., Liu, X., Yan, X., Gao, J., and Wei, F.
\newblock Augmenting language models with long-term memory, 2023{\natexlab{b}}.
\newblock URL \url{https://arxiv.org/abs/2306.07174}.

\bibitem[Wang \& Zhou(2024)Wang and Zhou]{wang2024chain}
Wang, X. and Zhou, D.
\newblock Chain-of-thought reasoning without prompting.
\newblock \emph{arXiv preprint arXiv:2402.10200}, 2024.

\bibitem[Wang et~al.(2023{\natexlab{c}})Wang, Wei, Schuurmans, Le, Chi, Narang, Chowdhery, and Zhou]{wang2023selfconsistency}
Wang, X., Wei, J., Schuurmans, D., Le, Q.~V., Chi, E.~H., Narang, S., Chowdhery, A., and Zhou, D.
\newblock Self-consistency improves chain of thought reasoning in language models.
\newblock In \emph{The Eleventh International Conference on Learning Representations}, 2023{\natexlab{c}}.
\newblock URL \url{https://openreview.net/forum?id=1PL1NIMMrw}.

\bibitem[Wang et~al.(2024{\natexlab{c}})]{wang-etal-2024-unleashing}
Wang, Z. et~al.
\newblock Unleashing the emergent cognitive synergy in large language models: A task-solving agent through multi-persona self-collaboration.
\newblock In \emph{Proceedings of the 2024 Conference of the North American Chapter of the Association for Computational Linguistics: Human Language Technologies (Volume 1: Long Papers)}, Jun. 2024{\natexlab{c}}.

\bibitem[Wang et~al.(2024{\natexlab{d}})]{wang2024largelanguagemodelenabled}
Wang, Z. et~al.
\newblock Large language model enabled semantic communication systems, 2024{\natexlab{d}}.

\bibitem[Wei et~al.(2022)Wei, Wang, Schuurmans, Bosma, Xia, Chi, Le, Zhou, et~al.]{wei2022chain}
Wei, J., Wang, X., Schuurmans, D., Bosma, M., Xia, F., Chi, E., Le, Q.~V., Zhou, D., et~al.
\newblock Chain-of-thought prompting elicits reasoning in large language models.
\newblock \emph{Advances in neural information processing systems}, 35:\penalty0 24824--24837, 2022.

\bibitem[Wu et~al.(2025)Wu, Zhu, Zhao, Yu, Ran, Wong, Sun, and Li]{wu_longattn_2025}
Wu, L., Zhu, D., Zhao, G., Yu, Z., Ran, J., Wong, X., Sun, L., and Li, S.
\newblock {LongAttn}: {Selecting} {Long}-context {Training} {Data} via {Token}-level {Attention}, February 2025.
\newblock URL \url{http://arxiv.org/abs/2502.16860}.
\newblock arXiv:2502.16860 [cs].

\bibitem[Wu et~al.(2024)]{wu2024autogen}
Wu, Q. et~al.
\newblock Autogen: Enabling next-gen {LLM} applications via multi-agent conversation, 2024.

\bibitem[Xiao et~al.(2024)Xiao, Tang, Zuo, Guo, Yang, Tang, Fu, and Han]{xiao_duoattention_2024}
Xiao, G., Tang, J., Zuo, J., Guo, J., Yang, S., Tang, H., Fu, Y., and Han, S.
\newblock {DuoAttention}: {Efficient} {Long}-{Context} {LLM} {Inference} with {Retrieval} and {Streaming} {Heads}, October 2024.
\newblock URL \url{http://arxiv.org/abs/2410.10819}.
\newblock arXiv:2410.10819 [cs].

\bibitem[Xu et~al.(2023)]{xu2023magic}
Xu, L. et~al.
\newblock Magic: Investigation of large language model powered multi-agent in cognition, adaptability, rationality and collaboration.
\newblock In \emph{ICLR 2024 Workshop on Large Language Model (LLM) Agents}, 2023.

\bibitem[Xu et~al.(2024)Xu, Ye, and Ren]{Xu2024StressTestingLL}
Xu, X., Ye, Q., and Ren, X.
\newblock Stress-testing long-context language models with lifelong icl and task haystack.
\newblock \emph{ArXiv}, abs/2407.16695, 2024.
\newblock URL \url{https://api.semanticscholar.org/CorpusID:271334664}.

\bibitem[Yang et~al.(2025)Yang, Yu, Li, Liu, Huang, Huang, Jiang, Tu, Zhang, Zhou, Lin, Dang, Yang, Yu, Li, Sun, Zhu, Men, He, Xu, Yin, Yu, Qiu, Ren, Yang, Li, Xu, and Zhang]{yang2025qwen251mtechnicalreport}
Yang, A., Yu, B., Li, C., Liu, D., Huang, F., Huang, H., Jiang, J., Tu, J., Zhang, J., Zhou, J., Lin, J., Dang, K., Yang, K., Yu, L., Li, M., Sun, M., Zhu, Q., Men, R., He, T., Xu, W., Yin, W., Yu, W., Qiu, X., Ren, X., Yang, X., Li, Y., Xu, Z., and Zhang, Z.
\newblock Qwen2.5-1m technical report, 2025.
\newblock URL \url{https://arxiv.org/abs/2501.15383}.

\bibitem[Yang et~al.(2018)Yang, Qi, Zhang, Bengio, Cohen, Salakhutdinov, and Manning]{yang-etal-2018-hotpotqa}
Yang, Z., Qi, P., Zhang, S., Bengio, Y., Cohen, W., Salakhutdinov, R., and Manning, C.~D.
\newblock {H}otpot{QA}: A dataset for diverse, explainable multi-hop question answering.
\newblock In Riloff, E., Chiang, D., Hockenmaier, J., and Tsujii, J. (eds.), \emph{Proceedings of the 2018 Conference on Empirical Methods in Natural Language Processing}, pp.\  2369--2380, Brussels, Belgium, October-November 2018. Association for Computational Linguistics.
\newblock \doi{10.18653/v1/D18-1259}.
\newblock URL \url{https://aclanthology.org/D18-1259}.

\bibitem[Yao et~al.(2023)Yao, Yu, Zhao, Shafran, Griffiths, Cao, and Narasimhan]{yao2023tree}
Yao, S., Yu, D., Zhao, J., Shafran, I., Griffiths, T.~L., Cao, Y., and Narasimhan, K.~R.
\newblock Tree of thoughts: Deliberate problem solving with large language models.
\newblock In \emph{Thirty-seventh Conference on Neural Information Processing Systems}, 2023.
\newblock URL \url{https://openreview.net/forum?id=5Xc1ecxO1h}.

\bibitem[Yin et~al.(2023)]{yin-etal-2023-exchange}
Yin, Z. et~al.
\newblock Exchange-of-thought: Enhancing large language model capabilities through cross-model communication.
\newblock In Bouamor, H., Pino, J., and Bali, K. (eds.), \emph{Proceedings of the 2023 Conference on Empirical Methods in Natural Language Processing}, pp.\  15135--15153, Singapore, December 2023. Association for Computational Linguistics.

\bibitem[Yuan et~al.(2025)Yuan, Gao, Dai, Luo, Zhao, Zhang, Xie, Wei, Wang, Xiao, Wang, Ruan, Zhang, Liang, and Zeng]{yuan_native_2025}
Yuan, J., Gao, H., Dai, D., Luo, J., Zhao, L., Zhang, Z., Xie, Z., Wei, Y.~X., Wang, L., Xiao, Z., Wang, Y., Ruan, C., Zhang, M., Liang, W., and Zeng, W.
\newblock Native {Sparse} {Attention}: {Hardware}-{Aligned} and {Natively} {Trainable} {Sparse} {Attention}, February 2025.
\newblock URL \url{http://arxiv.org/abs/2502.11089}.
\newblock arXiv:2502.11089 [cs].

\bibitem[Yuan et~al.(2024)Yuan, Ning, Zhou, Yang, Li, Zhuang, Tan, Yao, Lin, Li, Dai, Yan, and Wang]{yuan2024lveval}
Yuan, T., Ning, X., Zhou, D., Yang, Z., Li, S., Zhuang, M., Tan, Z., Yao, Z., Lin, D., Li, B., Dai, G., Yan, S., and Wang, Y.
\newblock Lv-eval: A balanced long-context benchmark with 5 length levels up to 256k, 2024.

\bibitem[Zhang et~al.(2024{\natexlab{a}})]{zhang-etal-2024-exploring}
Zhang, J. et~al.
\newblock Exploring collaboration mechanisms for {LLM} agents: A social psychology view.
\newblock In \emph{Proceedings of the Annual Meeting of the Association for Computational Linguistics}, Aug. 2024{\natexlab{a}}.

\bibitem[Zhang et~al.(2024{\natexlab{b}})Zhang, Chen, Hu, Xu, Chen, Hao, Han, Thai, Wang, Liu, and Sun]{zhang-etal-2024-bench}
Zhang, X., Chen, Y., Hu, S., Xu, Z., Chen, J., Hao, M., Han, X., Thai, Z., Wang, S., Liu, Z., and Sun, M.
\newblock $\infty${B}ench: Extending long context evaluation beyond 100{K} tokens.
\newblock In Ku, L.-W., Martins, A., and Srikumar, V. (eds.), \emph{Proceedings of the 62nd Annual Meeting of the Association for Computational Linguistics (Volume 1: Long Papers)}, pp.\  15262--15277, Bangkok, Thailand, August 2024{\natexlab{b}}. Association for Computational Linguistics.
\newblock URL \url{https://aclanthology.org/2024.acl-long.814}.

\bibitem[Zhang et~al.(2024{\natexlab{c}})Zhang, Khalifa, Logeswaran, Kim, Lee, Lee, and Wang]{zhang-etal-2024-small}
Zhang, Y., Khalifa, M., Logeswaran, L., Kim, J., Lee, M., Lee, H., and Wang, L.
\newblock Small language models need strong verifiers to self-correct reasoning.
\newblock In Ku, L.-W., Martins, A., and Srikumar, V. (eds.), \emph{Findings of the Association for Computational Linguistics: ACL 2024}, pp.\  15637--15653, Bangkok, Thailand, August 2024{\natexlab{c}}. Association for Computational Linguistics.
\newblock \doi{10.18653/v1/2024.findings-acl.924}.
\newblock URL \url{https://aclanthology.org/2024.findings-acl.924/}.

\bibitem[Zhang et~al.(2024{\natexlab{d}})Zhang, Sun, Chen, Pfister, Zhang, and Arik]{zhang2024chainagentslargelanguage}
Zhang, Y., Sun, R., Chen, Y., Pfister, T., Zhang, R., and Arik, S.~O.
\newblock Chain of agents: Large language models collaborating on long-context tasks, 2024{\natexlab{d}}.
\newblock URL \url{https://arxiv.org/abs/2406.02818}.

\bibitem[Zhao et~al.(2024{\natexlab{a}})Zhao, Zu, Xu, Lu, He, Ding, Gui, Zhang, and Huang]{zhao2024longagentscalinglanguagemodels}
Zhao, J., Zu, C., Xu, H., Lu, Y., He, W., Ding, Y., Gui, T., Zhang, Q., and Huang, X.
\newblock Longagent: Scaling language models to 128k context through multi-agent collaboration, 2024{\natexlab{a}}.
\newblock URL \url{https://arxiv.org/abs/2402.11550}.

\bibitem[Zhao et~al.(2024{\natexlab{b}})]{zhao2024competeai}
Zhao, Q. et~al.
\newblock Compete{AI}: Understanding the competition dynamics of large language model-based agents.
\newblock In \emph{Agentic Markets Workshop at ICML 2024}, 2024{\natexlab{b}}.

\bibitem[Zhao et~al.(2024{\natexlab{c}})Zhao, Zhang, Pan, Yao, Yu, Wu, and Chen]{zhao-etal-2024-fact}
Zhao, X., Zhang, H., Pan, X., Yao, W., Yu, D., Wu, T., and Chen, J.
\newblock Fact-and-reflection ({F}a{R}) improves confidence calibration of large language models.
\newblock In Ku, L.-W., Martins, A., and Srikumar, V. (eds.), \emph{Findings of the Association for Computational Linguistics: ACL 2024}, pp.\  8702--8718, Bangkok, Thailand, August 2024{\natexlab{c}}. Association for Computational Linguistics.
\newblock \doi{10.18653/v1/2024.findings-acl.515}.
\newblock URL \url{https://aclanthology.org/2024.findings-acl.515/}.

\bibitem[Zhao et~al.(2024{\natexlab{d}})]{10531769}
Zhao, Y. et~al.
\newblock Lamosc: Large language model-driven semantic communication system for visual transmission.
\newblock \emph{IEEE Transactions on Cognitive Communications and Networking}, 10\penalty0 (6):\penalty0 2005--2018, 2024{\natexlab{d}}.

\bibitem[Zhong et~al.(2021)Zhong, Yin, Yu, Zaidi, Mutuma, Jha, Awadallah, Celikyilmaz, Liu, Qiu, and Radev]{zhong-etal-2021-qmsum}
Zhong, M., Yin, D., Yu, T., Zaidi, A., Mutuma, M., Jha, R., Awadallah, A.~H., Celikyilmaz, A., Liu, Y., Qiu, X., and Radev, D.
\newblock {QMS}um: A new benchmark for query-based multi-domain meeting summarization.
\newblock In Toutanova, K., Rumshisky, A., Zettlemoyer, L., Hakkani-Tur, D., Beltagy, I., Bethard, S., Cotterell, R., Chakraborty, T., and Zhou, Y. (eds.), \emph{Proceedings of the 2021 Conference of the North American Chapter of the Association for Computational Linguistics: Human Language Technologies}, pp.\  5905--5921, Online, June 2021. Association for Computational Linguistics.
\newblock \doi{10.18653/v1/2021.naacl-main.472}.
\newblock URL \url{https://aclanthology.org/2021.naacl-main.472/}.

\bibitem[Zhou et~al.(2023)Zhou, Sch{"a}rli, Hou, Wei, Scales, Wang, Schuurmans, Cui, Bousquet, Le, and Chi]{zhou2023leasttomost}
Zhou, D., Sch{"a}rli, N., Hou, L., Wei, J., Scales, N., Wang, X., Schuurmans, D., Cui, C., Bousquet, O., Le, Q.~V., and Chi, E.~H.
\newblock Least-to-most prompting enables complex reasoning in large language models.
\newblock In \emph{The Eleventh International Conference on Learning Representations}, 2023.
\newblock URL \url{https://openreview.net/forum?id=WZH7099tgfM}.

\bibitem[Zhou et~al.(2024)Zhou, Pujara, Ren, Chen, Cheng, Le, Chi, Zhou, Mishra, and Zheng]{zhou2024selfdiscover}
Zhou, P., Pujara, J., Ren, X., Chen, X., Cheng, H.-T., Le, Q.~V., Chi, E.~H., Zhou, D., Mishra, S., and Zheng, S.
\newblock {SELF}-{DISCOVER}: Large language models self-compose reasoning structures.
\newblock In \emph{The Thirty-eighth Annual Conference on Neural Information Processing Systems}, 2024.
\newblock URL \url{https://openreview.net/forum?id=BROvXhmzYK}.

\bibitem[Zhou et~al.(2025)Zhou, Yin, Zuo, and Cheng]{zhou_progressive_2025}
Zhou, Q., Yin, P., Zuo, P., and Cheng, J.
\newblock Progressive {Sparse} {Attention}: {Algorithm} and {System} {Co}-design for {Efficient} {Attention} in {LLM} {Serving}, March 2025.
\newblock URL \url{http://arxiv.org/abs/2503.00392}.
\newblock arXiv:2503.00392 [cs].

\bibitem[Zhuang et~al.(2024)Zhuang, Zhang, and Hu]{Zhuang2024PoSE}
Zhuang, B., Zhang, C., and Hu, Z.
\newblock Pose: Suppressing perceptual noise in embodied agents for enhanced semantic navigation.
\newblock \emph{IEEE Robotics and Automation Letters}, 9:\penalty0 963--970, 2024.

\end{thebibliography}

\newpage
\appendix
\onecolumn
\section{Proof of Completeness}
\label{sec:proof}
\textbf{Hypothetical Premise}: the LLM($I$,$\theta$) acting an agent can complete the problem decomposition and reasoning of a specified context chunk within k tokens context window. 

\textbf{Problem Definition}: Reasoning based on information distributed in different chunks has a certain dependency sequence. If chunks are not processed in this sequence, it may lead to the breakage of the reasoning process. This sequential dependency between reasoning across different chunks can be reduced to a topological sorting problem of predecessor-successor relationships in an \textbf{AOV (Activity On Vertex) network}. In the scenario of long text reasoning, the dependency order of chunks can change along with the variation of the entities of concern, which can be expressed as:
\begin{equation}
M = 
\begin{pmatrix}
m_{1,1} & m_{1,2} & \cdots & m_{1,y} \\
m_{2,1} & m_{2,2} & \cdots & m_{2,y} \\
\vdots  & \vdots  & \ddots & \vdots  \\
m_{x,1} & m_{x,2} & \cdots & m_{x,y}
\end{pmatrix}
\end{equation}
where $m_{x,y}$ denotes the the $y^{th}$ chunk should be dealt with on the ${m_{x,y}}^{th}$ one in the topological sequence. For example, $\{4,2,1,3\}$ means reasoning should be done one-by-one on chunk 3, 2, 4, 1 successively.

Explorer scans through the chunks in one direction and Decider proposes replay when reaching the last column. The next starting point is set at the chunk, denoted as $o^{th}$ chunk, where the last unanswered problem $u$ is proposed. \textbf{The starting points of different rows are the same}. 
\begin{equation}
o = \max_{i \in [1,x]} \left( \arg\max_j u_{i,j} \right)
\end{equation}
The Explorer accepts $m_{x,y}$=1 and then 2, and so on until all chunks are received. Specifically, in ($y-1$) rounds of scan direction reverse, the Explorer would accept all items in M. 

For the first starting point of reverse:
\begin{equation}
    \forall i \in [1,x],\forall j > o, m_{i,j} \neq m_{i,o}+1
\end{equation}
\begin{equation}
    \forall i \in [1,x], \exists j_0 \in [1,o], m_{i,j_0} = m_{i,o}+1 
\end{equation}
The starting point of the second round of reverse can be determined by:
\begin{equation}
    o = \min_{i \in [1,x]} \left( \arg\min_j u_{i,j} \right)
\end{equation}
For the second starting point of reverse:
\begin{equation}
    \forall i \in [1,x],\forall j < o, m_{i,j} \neq m_{i,o}+1
\end{equation}
\begin{equation}
    \forall i \in [1,x], \exists j_0 \in [o,y], m_{i,j_0} = m_{i,o}+1 
\end{equation}
The two processes are mirrored,
\begin{equation}
M^{mir} = 
\begin{pmatrix}
m_{1,y} & m_{1,y-1} & \cdots & m_{1,1}\\
m_{2,y} & m_{2,y-1} & \cdots & m_{2,1}\\
\vdots  & \vdots  & \ddots & \vdots\\
m_{x,y} & m_{x,y-1} & \cdots & m_{x,1}
\end{pmatrix}
\end{equation}
Augment the matrix by concatenate $M$ and $M^{mir}$, 
\begin{equation}
    \widetilde{M} = \Big[ M \Big| M^\text{mir} \Big| M \Big| M^\text{mir} \Big| \cdots \Big]_{x \times y^2}.
\end{equation}
Scan from left to right. For every y chunks, at least one chunk is accepted. So all items in M can be accepted in $y$ scan, namely $y-1$ replay. It's not non-trivial, for:
\begin{equation}
    \forall x, XpandA\ accept\ M.
\end{equation}

\end{document}